\documentclass{article}

\usepackage[final]{neurips_2025}

\usepackage[utf8]{inputenc}
\usepackage[T1]{fontenc}
\usepackage{hyperref}
\usepackage{url}
\usepackage{booktabs}
\usepackage{amsfonts}
\usepackage{nicefrac}
\usepackage{microtype}
\usepackage{xcolor}
\usepackage{amsmath}
\usepackage{graphicx}
\usepackage{amsthm}
\usepackage{amssymb}
\usepackage{algorithm}
\usepackage{algorithmic}
\usepackage{subcaption}
\usepackage[inline]{enumitem}
\usepackage{tabularx}
\usepackage{lscape}
\title{Off-policy Reinforcement Learning with Model-based Exploration Augmentation}

\author{
Likun Wang$^{1 \dagger}$ \quad Xiangteng Zhang$^{1 \dagger}$ \quad Yinuo Wang$^{1 \dagger}$ \quad Guojian Zhan$^1$ \\
\textbf{Wenxuan Wang$^1$ \quad Haoyu Gao$^1$ } \quad
\textbf{Jingliang Duan$^{1,2*}$ \quad Shengbo Eben Li$^{1}\thanks{Corresponding author <duanjl15@163.com> <lishbo@tsinghua.edu.cn>.}$} \vspace{0.15cm}\\
$^1$School of Vehicle and Mobility \&  College of AI, Tsinghua University \\ $^2$School of Mechanical Engineering, University of Science and Technology Beijing
}

\newtheorem{theorem}{Theorem}
\newtheorem{lemma}{Lemma}

\begin{document}

\maketitle

\begin{abstract}

Exploration is fundamental to reinforcement learning (RL), as it determines how effectively an agent discovers and exploits the underlying structure of its environment to achieve optimal performance.
Existing exploration methods generally fall into two categories: active exploration and passive exploration.
The former introduces stochasticity into the policy but struggles in high-dimensional environments, while the latter adaptively prioritizes transitions in the replay buffer to enhance exploration, yet remains constrained by limited sample diversity.
To address the limitation in passive exploration, we propose \textbf{Mo}delic \textbf{G}enerative \textbf{E}xploration (MoGE), which augments exploration through the generation of under-explored critical states and synthesis of dynamics-consistent experiences through transition models.
MoGE is composed of two components: (1) a diffusion-based generator that synthesizes critical states under the guidance of a utility function evaluating each state’s potential influence on policy exploration, and (2) a one-step imagination world model for constructing critical transitions based on the critical states for agent learning.
Our method adopts a modular formulation that aligns with the principles of off-policy learning, allowing seamless integration with existing algorithms to improve exploration without altering their core structures.
Empirical results on OpenAI Gym and DeepMind Control Suite reveal that MoGE effectively bridges exploration and policy learning, leading to remarkable gains in both sample efficiency and performance across complex control tasks.
\end{abstract}

\section{Introduction}

Reinforcement learning has demonstrated remarkable potential across various tasks, including autonomous driving, large language models, game playing, and embodied artificial intelligence \cite{chen2021interpretable,dogru2024reinforcement,wang2024reinforcement,schrittwieser2020mastering,gupta2021embodied,wang2024diffusion, wang2025enhanced}. It optimizes policies through trial and error, with policy performance fundamentally relying on the diversity and coverage of samples collected during interaction with environments. \cite{grondman2012survey,mnih2016asynchronous,haarnoja2018soft}.
Similar to imitation learning (IL), RL algorithms face out-of-distribution (OOD) challenges due to limited diversity in training data. However, RL mitigates this issue by leveraging its exploration capabilities to reach unvisited regions of the state space. In this context, enhancing exploration strategies to collect diverse samples and achieve broader state space coverage becomes crucial for improving the effectiveness and generalization of RL algorithms \cite{han2021diversity,banerjee2023boosting}.

Conventional exploration strategies can be broadly classified into two categories: active exploration purely based on policy, and passive exploration based on states.
Existing approaches achieve active exploration by introducing randomness into the policy \cite{plappert2017parameter,fujimoto2018addressing} or adding exploration bonuses, which help prevent the policy from converging prematurely to a narrow subset of actions \cite{ahmed2019understanding}.
Specifically, methods like SAC \cite{haarnoja2018soft} and DSAC \cite{duanDistributional2025} leverage maximum entropy as exploration bonuses to encourage exploration, while algorithms such as MPO \cite{abdolmaleki2018max} employ multimodal policies to maintain action diversity and stochasticity.
However, the active exploration mechanisms are inherently limited by the agent's actual interacted trajectories, which are constrained by the environment's initial states, finite episode lengths, and the agent's current policy \cite{varakantham2023constrained}. As a result, many critical regions distant from typical rollouts or rarely visited often remain unexplored, reinforcing value estimation errors and narrowing the scope of policy learning \cite{li2023near,flennerhag2020temporal}. Furthermore, forcing exploration through the policy can divert focus from reward optimization, leading to suboptimal learning \cite{chen2022redeeming}.

In contrast to active exploration, which relies on policy-driven interactions, passive exploration modifies the sample distribution of the agent to incorporate prioritized information.
Passive exploration originates from Prioritized Experience Replay (PER) \cite{shin2017continual}, which enhances learning by selectively reusing valuable experiences but remains fundamentally limited to previously collected samples. To address this restriction, certain approaches like SER \cite{lu2023synthetic} and PGR \cite{wang2024prioritized} leverage generative models to augment the replay buffer, thereby artificially intensifying the state-action distribution. However, this expansion is still confined to the vicinity of observed data, adhering closely to the original data distribution. Other methods integrate world models with current policies to simulate state transitions, potentially generating higher-quality transitions \cite{hafner2019dream,hafner2020mastering,hafner2023mastering}; nevertheless, these generated samples typically exhibit limited diversity due to their strong dependence on the policy-conditioned transitions originating from existing states. Moreover, generating complete transitions in high-dimensional state spaces further amplifies the bias introduced by synthetic data, as the complexity grows exponentially with dimensionality \cite{wang2025deep}.

To tackle the issues above in policy exploration, we propose MoGE, a novel exploration paradigm for enhancing off-policy RL algorithms by generating critical transitions across the entire state space guided by exploratory priors.
It consists of two components: a generator that produces critical states with high exploratory potential for the current policy and value network, and a dynamics model that simulates one-step transitions, enabling the generation of critical transitions.
Specifically, our work makes three main contributions:
\textbf{(1)}
We employ a conditional diffusion generator to sample critical states with high exploratory potential. To guarantee the state-space compliance and feasibility of the generated states, we theoretically prove that the state distribution in the replay buffer asymptotically converges to the stationary occupancy measure of the optimal policy. By continuously fine-tuning the generator on the replay buffer, we ensure that its learned distribution shares a common support with the optimal policy's occupancy measure, generating critical states in a compliant way.
\textbf{(2)}

To guarantee the dynamical consistency of the generated samples, we design a one-step imagination world model to imitate the dynamics of the environment. This world model allows for efficient pre-training through supervised learning, supporting the construction of training experiences and designing the classifier of the conditional diffusion-based critical state generator.
\textbf{(3)}
We propose an off-policy RL training framework that integrates MoGE seamlessly into existing algorithms without requiring any modifications to their original structure. By introducing importance sampling that mixes critical transitions generated by MoGE with replay buffer samples, MoGE enhances exploration, leading to improved performance and sample efficiency.
Experiments on standard continuous control benchmarks, including OpenAI Gym \cite{brockman2016openai} and DeepMind Control Suite \cite{tassa2018deepmind} demonstrate that MoGE, as a plug-in module, consistently improves both the final performance and the sample efficiency of baseline off-policy RL algorithms.
\section{Preliminaries}
\subsection{Reinforcement Learning and Policy Exploration}

In RL, an agent interacts with an environment modeled as a Markov Decision Process (MDP) \cite{sutton1998reinforcement}, defined as $M = (\mathcal{S}, \mathcal{A}, P, r, \gamma)$, where $\mathcal{S}$ and $\mathcal{A}$ represent the state and action spaces, $P: \mathcal{S} \times \mathcal{A} \to \mathcal{S}$ denotes the environment's transition dynamics, $r: \mathcal{S} \times \mathcal{A} \to \mathbb{R}$ is the reward function, and $\gamma \in [0, 1)$ is the discount factor \cite{li2023reinforcement}.

To formally describe the target of RL, we introduce the \textit{occupancy measure} $d^\pi(s)$, which is defined as $d^\pi(s) = (1 - \gamma) \sum_{t=0}^\infty \gamma^t \mathbb{P}(s_t = s \mid \pi)$. This occupancy measure represents the visitation frequency of a given state $s$ under the policy $\pi$ \cite{laroche2023occupancy}. Likewise, we can define the $d^\pi(s, a) = d^\pi(s)\pi(a|s)$, which represents the visitation frequency of a given state-action pair $(s, a)$.
The target of policy is to maximize the expected cumulative reward, which can be expressed in terms of the occupancy measure as $J(\pi) = \mathbb{E}_{(s, a) \sim d^\pi(s, a)}[r(s, a)]$. Meanwhile, a value function is trained to estimates the value of the current policy by minimizing the temporal difference (TD) error, which is formulated as $J(\pi)=\mathbb{E}_{(s, a) \sim d^\pi(s, a)} \left[\left(Q^\pi(s, a) - (r + \gamma Q^\pi(s', a'))\right)^2\right]$, where $Q^\pi(s, a)$ denote the value function that evaluate the quality of given $(s, a)$. Through this formulation, the actor optimizes the policy towards higher cumulative rewards, while the critic evaluates and stabilizes the learning process by minimizing prediction errors.

Effective policy training depends on discovering policy-improving states. Since such critical states often require active exploration, identifying them under the current policy enables targeted updates that accelerate learning. If these states can be estimated, they can be selectively replayed for more efficient updates. In existing RL frameworks, two common metrics are used to quantify their criticality:

\textbf{Policy entropy}.
The policy entropy reflects the randomness of action selection at a given state.
High entropy may indicate either insufficient visitation, offering high information gain, or proximity to critical decision points in the MDP where small action changes lead to divergent outcomes \cite{neu2017unified,eysenbach2019if}. Focusing learning on such regions enhances policy robustness and long-horizon decision-making. For example, in the case of a Gaussian distribution, the utility function is defined as:
\begin{equation}
    f(s) = \mathcal{H}(\pi(\cdot|s)) = \frac{1}{2} \log\bigl((2\pi e)^d \det \Sigma(s)\bigr),
\end{equation}
where $\pi(\cdot|s)=N(\cdot;u(s),\Sigma(s))$ while $\pi$ and $e$ denote the constant.

\textbf{TD error}. High TD error states are critical for exploration, as they highlight regions where the value function poorly approximates returns under the current policy \cite{janz2019successor,flennerhag2020temporal}. These states often correspond to areas of high uncertainty or insufficient optimization. Prioritizing them helps reduce value bias, correct suboptimal actions, and improve policy robustness. The TD error for a given state under the current policy can be estimated as:
\begin{equation}
f(s)=\delta_t(s) \approx r(s,\pi_{\theta}(s)) + \gamma Q_\psi(d_{\phi}(s,\pi_{\theta}(s)), \pi(d_{\phi}(s,\pi_{\theta}(s)))) - Q_\psi(s, \pi_{\theta}(s)).
\label{eq:td}
\end{equation}

\subsection{Diffusion Models for Generative Tasks}
Diffusion models have emerged as effective generative models due to their ability to capture complex data distributions \cite{ho2020denoising,sohl2015deep,song2019generative}. Inspired by non-equilibrium thermodynamics, they simulate two complementary processes: a forward diffusion process incrementally adding noise to the original data, and a reverse denoising process reconstructing the data distribution from noise.
In the forward diffusion process, an initial data sample $s_0 \sim q(s_0)$ is gradually transformed into Gaussian noise $s_T$ by iteratively applying:
\begin{equation}
    q(s_t|s_{t-1}) = \mathcal{N}(s_t;\sqrt{1-\beta_t}s_{t-1},\beta_t\mathbf{I}),
\end{equation}
where $\beta_t$ controls the noise schedule. Letting $\alpha_t = 1 - \beta_t$ and $\bar{\alpha}_t = \prod_{i=1}^{t} \alpha_i$, the true reverse distribution conditioned on $s_0$ is given by:
\begin{equation}
    q(s_{t-1}|s_t, s_0) = \mathcal{N}\left(s_{t-1}; \frac{1}{\sqrt{\alpha_t}}\left(s_t - \frac{1-\alpha_t}{\sqrt{1-\bar{\alpha}_t}}\epsilon_t\right), \frac{1-\bar{\alpha}_{t-1}}{1-\bar{\alpha}_t}\beta_t\mathbf{I}\right).
\end{equation}

Since $s_0$ is unknown during generation, diffusion models approximate this posterior with a parameterized neural network $\epsilon_\varphi(s_t, t)$ to predict the noise term $\epsilon_t$. To align with the real reverse process, the approximate reverse process is expressed as:
\begin{equation}
    p_\varphi(s_{t-1}|s_t) = \mathcal{N}\left(s_{t-1}; \frac{1}{\sqrt{\alpha_t}}\left(s_t - \frac{\beta_t}{\sqrt{1-\bar{\alpha}_t}}\epsilon_\varphi(s_t, t)\right), \frac{1-\bar{\alpha}_{t-1}}{1-\bar{\alpha}_t}\beta_t\mathbf{I}\right).
\end{equation}

The training objective is to minimize the noise prediction error:
\begin{equation}
    \mathcal{L}_{\text{generator}} = \mathbb{E}_{s_0, \epsilon, t}\left[\|\epsilon - \epsilon_\varphi(\sqrt{\bar{\alpha}_t} s_0 + \sqrt{1-\bar{\alpha}_t}\epsilon, t)\|_2^2\right], \quad \epsilon \sim \mathcal{N}(0, \mathbf{I}),
    \label{eq: generation loss}
\end{equation}
enabling efficient data generation by progressively denoising from standard Gaussian noise.

\section{Methods}

\begin{figure}[t]
  \centering
    \includegraphics[width=\textwidth,trim={0.28cm 0.25cm 0.28cm 0.25cm},   clip]{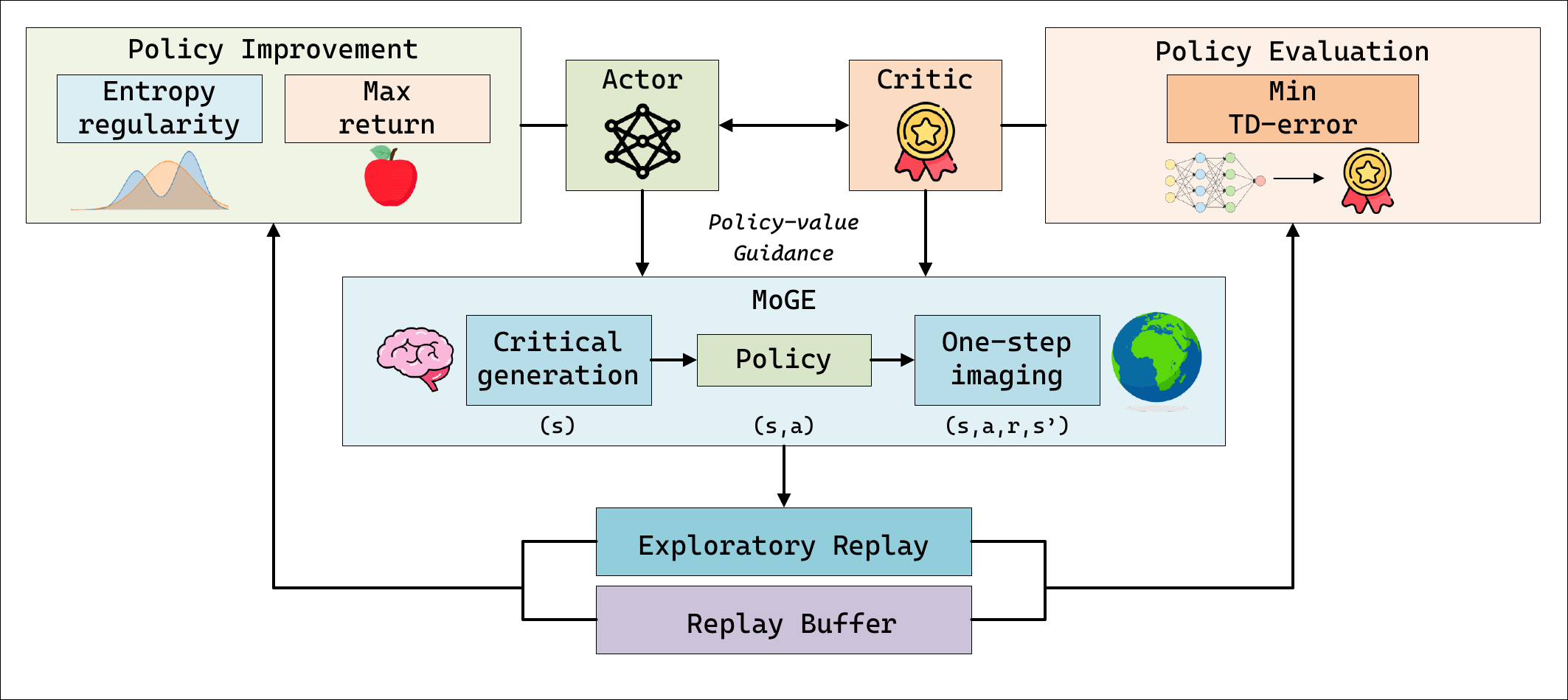}
  \caption{\textbf{Overview of MoGE.} MoGE is composed of two sub-modules: a generator and a one-step world model. The generator produces critical states under-explored but potentially valuable for policy exploration under the guidance of policy and value function, while the one-step world model predicts the next state and reward to construct the transitions. The formulated exploratory can be mixed with real samples from the buffer to perform the policy improvement and evaluation.}
  \label{fig:method}
\end{figure}

In this section, we introduce the MoGE paradigm, illustrated in Figure \ref{fig:method}.  In Section \ref{sec:ctitical}, we focus on the method and theoretical foundation of generating critical states through the policy model, as well as the selection of guidance methods. In Section \ref{sec:wm}, we introduce the structure and application of the one-step imagination world model. In Section \ref{sec: quality analyses}, we first analyze the quality of samples generated by MoGE in both novelty and dynamics consistency, and after that, we propose a training framework that integrates MoGE with existing off-policy actor-critic methods in Section \ref{sec:mgpe}.

\subsection{Critical State Generation with Steady Occupancy Measurement Alignment}
\label{sec:ctitical}
State-based passive exploration leverages prior knowledge to uncover policy-improving states without active search, enabling more directed and sample-efficient policy training.
However, when such states are not visited by the policy or stored in the replay buffer, they cannot be utilized for learning. Generative models overcome this limitation by enabling the synthesis of specific states once the model is trained.
By leveraging advanced generative techniques such as diffusion and flow matching \cite{lipman2022flow,geng2025mean}, the model is capable of synthesizing high-fidelity samples that accurately approximate even intractable target distributions.

Motivated by the insight above, MoGE adopts a classifier-guided diffusion model \cite{dhariwal2021diffusion}, $p_{\varphi}(s|c)$, which serves as a utility-conditioned generator to synthesize policy-improving states.
The unconditional diffusion model learns the manifold of plausible states, while a separately trained classifier provides directional guidance, steering generation toward high-utility regions. This guided synthesis bridges generative modeling and policy optimization, allowing the agent to access critical yet previously unvisited states.
In this classifier-guided diffusion framework, the gradient used during inference combines the unconditional diffusion gradient with a classifier-based gradient. Specifically, implementing Bayes' formulation and denoting the classifier $c$ by a continuous \textit{utility function} $f(s)$, the classifier-guided gradient can be expressed as:

\begin{equation}
\nabla\mathrm{log}p_{\varphi}(s_t|f(s_t))=\underbrace{\nabla\mathrm{log}p_{\varphi}(s_t)}_{\textbf{Diffusion gradient}}+\underbrace{\omega\nabla\mathrm{log}p(f(s_t)|s_t)}_\textbf{Classifier utility gradient},
\end{equation}

where $\omega$ is the guidance scale, which controls the strength of this guidance, balancing between following the unconditional diffusion prior and aligning with the target utility function. By selecting an appropriate utility function $f(s)$, this approach allows targeted control over state generation probability, enabling effective exploration of high-value regions.

However, a central challenge in training generative models within RL settings lies in determining an appropriate stationary target distribution. Since the data distribution in RL is inherently policy-dependent and evolves as the policy updates, obtaining a stable and realistic state distribution becomes infeasible, thereby complicating the training of the generator.
As a result, arbitrary or poorly aligned state generation can lead to exploration inefficiencies and instability during policy updates.
To address this, the generated states must align closely with an approximately physically grounded and stable occupancy measure, ensuring both state-space compliance and dynamical consistency.
In practice, we observe that the behavior policy in the replay buffer exhibits statistically diminishing variation over training, inducing an approximately stationary occupancy measure that can serve as a physically grounded target for the generative model.
To formally establish this alignment, we first propose the following theorem as a theoretical guarantee.

\begin{theorem}[Steady‑State Occupancy Measurement Alignment Theorem]
Let $\beta(a \mid s)$ be a behavior policy, $\pi^{*}(a \mid s)$ be a specific static policy, and let $\nu_{t}(s)$ and $d^{\pi^{*}}(s)$ represent the state occupancy measures under $\beta(a \mid s)$ and $\pi^*(a \mid s)$, respectively.
Assuming that the divergence between the policy and $\pi^{*}(s)$ gradually decreases over the course of training.
and that the replay buffer has finite capacity following a First-In-First-Out (FIFO) scheme with states sampled exclusively from the current policy interaction, the following convergence relationship holds:
\begin{equation}
\lim_{t \to \infty} \operatorname{TV}(\nu_t(s), d^{\pi^{*}}(s)) = 0.
\end{equation}
\end{theorem}

\begin{proof}
See Appendix \ref{proof 1}.
\end{proof}

By aligning the unconditional diffusion model's training distribution with the stable occupancy measure in the replay buffer, the critical state generator inherently guarantees generated states remain valid within the true state space since the conditional diffusion shares the same support with the unconditional one \cite{dhariwal2021diffusion,song2020score}.
This alignment ensures high-quality, stable critical state generation and enhances the reliability and efficacy of subsequent policy improvements.
In this work, we introduce two utility functions below that satisfy these requirements and analyze how each facilitates policy exploration and learning. Since the MoGE is equipped with a transition model, which enables the computation and gradient propagation of certain utility functions that would otherwise be intractable. The two criteria are \textbf{policy entropy} and \textbf{TD error} (see preliminaries for a detailed introduction).
During training, we use different utility signals for different tasks, which facilitates more effective policy learning and accelerates value function convergence.

\subsection{Transition Imaging with One-step World Model}
\label{sec:wm}
When the environment can not return the reward $r$ and next state $s'$ for arbitrary $(s, a)$, the utility function $U$ cannot be directly evaluated without environment interaction.
Therefore, we employ learned surrogate models---a reward model $r_{\psi}(s, a)$ and a dynamics model $f_{\theta}(s, a)$---to approximate these quantities and provide differentiable estimates of $r$ and $s'$. To work in conjunction with the critical state generator, we introduce a one-step imagination world model $\mathcal{F}_\phi$ to estimate the environment dynamics.
Since we need to estimate the environment’s transition and reward functions for arbitrary $(s, a)$ pairs, we only require the world model to be accurate for one-step dynamics under the current state and policy.
Consequently, our world model diverges from conventional designs \cite{hafner2019dream} in both structure and training approach. By focusing on one-step predictions, it trades off long-horizon accuracy to ensure reliable transitions within the region characterized by the optimal policy’s occupancy measure.

Our proposed one-step imagination world model $\mathcal{F}_\phi$ consists of five parameterized components to predict the following variables:
\begin{equation}
\begin{array}{l@{\quad}l}
\text{Representation:} & \mathbf{z}_t = g_\phi(s_t) \\
\text{Reconstruction:} & s_t = h_\phi(\mathbf{z}_t) \\
\text{Latent dynamics:} & \mathbf{z}_{t+1} = d_\phi(\mathbf{z}_t, a_t) \\
\text{Reward:} & \hat{r}_t = R_\phi(\mathbf{z}_t, a_t) \\
\text{Termination prediction:} & \hat{c}_t = C_\phi(\mathbf{z}_t, a_t). \\

\end{array}
\end{equation}

The detailed structure of the world model is depicted in Figure \ref{fig:wm} of Appendix \ref{wm}.
Departing from prior work that relies on probabilistic components \cite{hafner2020mastering}, we find that implementing all modules of the world model as deterministic networks suffices for effective performance. Except for the latent dynamics model, which uses a two-layer \textit{Transformer} encoder \cite{vaswani2017attention} to flexibly handle both sequential and single-step inputs, all other components are implemented as standard MLPs.
At each time step $ t $, the observation $ s_t $ is first encoded into a latent representation $ \mathbf{z}_t $ using an encoder network $ g_\phi $. Given latent state  $ \mathbf{z}_t $ in feature space and the executed action $ a_t $, the world model outputs:
$\textbf{(1)}$ a prediction of the next latent state $ \mathbf{z}_{t+1} $,
$\textbf{(2)}$ the corresponding one-step reward $\hat{r}_t$ and $\textbf{(3)}$ the termination factor that evaluates whether the transition is terminated.

During training, we treat the reward signal and termination factor as augmentations of the next-step state, and design separate loss functions through three output heads. This design simplifies the architecture while preserving the predictive capacity of the world model. The total training loss of the world model is presented as:

\begin{equation}
\begin{aligned}
\mathcal{L}_\text{worldmodel}(\phi) = \frac{1}{BT}\sum_{n=1}^B\sum_{t=1}^T \Big[
& \underbrace{||\hat{s}_t - s_t||_2}_{\textbf{reconstruction}} +
\underbrace{||\hat{r}_t - r_t||_2}_{\textbf{reward}} +
\underbrace{c_t \log \hat{c}_t + (1 - c_t) \log(1 - \hat{c}_t)}_{\textbf{termination}} \\
& + \underbrace{\beta_1 ||\textbf{sg}(g_{\phi}(s_{t+1})) - d_{\phi}(g_{\phi}(s_{t}), a_t)||_2}_{\textbf{dynamics}}  \\
& + \underbrace{\beta_2 ||g_{\phi}(s_{t+1}) - \textbf{sg}(d_{\phi}(g_{\phi}(s_{t}), a_t))||_2}_{\textbf{representation}} s
\Big],
\end{aligned}
\label{eq:worldmodel}
\end{equation}
where $\beta_1=0.5,\beta_2=0.1$,$(s_t,a_t,r_t,s_{t+1})_{0:T}$ are the transitions sampled from buffer $\mathcal{B}$, and $\textbf{sg}(\cdot)$ is the stop-gradient operator. $T$ is the total length of the transition that depends on the sample method, and ${B}$ is the batch size. The model accommodates both sequence and single-step inputs, with
$T=1$ indicates the single-step setting.
\begin{algorithm}[t]
\caption{Off-policy RL training framework with MoGE}
\label{alg:mac}
\textbf{Input:} Policy $\pi_\theta$, critical-state generator $p_\varphi$, One-step world model $\mathcal{F}_\phi$, critic $Q_{\psi}$ \\

\textbf{Initialize:} $\pi_\theta$, $p_\varphi$, $\mathcal{F}_\phi$, $Q_{\psi}$
\begin{algorithmic}[1]
    \STATE \texttt{// Pretraining}
    \STATE Interact with environment using random actions: $(r, s', c) \leftarrow \texttt{env.step}(\text{random})$
    \STATE Store random transitions $(s, a_\text{rand}, r, s')$ into replay buffer $\mathcal{B}$
    \STATE Update $\mathcal{F}_\phi$, by minimizing model loss $\mathcal{L}_{\text{worldmodel}}$
    in \eqref{eq:worldmodel} using random interaction data
    \STATE Re-warmup the buffer $\mathcal{B}$ with the current target $\pi_\theta$
    \FOR{each iteration}
        \STATE \texttt{// Data Collection}
        \STATE Initialize $s$
        \STATE Interact with environment: $(r, s', c) \leftarrow \texttt{env.step}(\pi_{\theta}(s))$
        \STATE Store transition $(s, \pi_{\theta}(s), r, s')$ into replay buffer $\mathcal{B}$
        \STATE \texttt{// Off-policy RL Learning}
        \STATE Sample transitions $\Gamma=\{(s_t, a_t, r_t, s_{t+1})_{t=0}^T\} \sim \mathcal{B}$

        \STATE Update $\mathcal{F}_\phi$,$p_\varphi$ by minimizing model loss $\mathcal{L}_{\text{worldmodel}}$ and $\mathcal{L}_{\text{generator}}$ in \eqref{eq:worldmodel},\eqref{eq: generation loss} using $\Gamma$
        \STATE Generate critical states $s_{e}$ by using classifier-guidance generator$p_\varphi$ with utility function $f(s)$
        \STATE Generate full transitions with one-step imaging: $(r, s'_{e}, c) \leftarrow \mathcal{F}_{\phi}(s_{e},\pi_{\theta}(s_{e}))$
        \STATE Combine critical transitions $\Gamma_{e}=\{(s_{e}, \pi_{\theta}(s_{e}), r, s'_{e})\}$ with $\Gamma$ : $\Gamma'=\Gamma \cup\Gamma_{e}$

        \STATE Update $\pi_\theta$, $Q_{\psi}$ with downstream algorithm using $\Gamma'$ (for $\pi_\theta$, do approximate importance sampling with $\lambda$-mixture )
    \ENDFOR
\end{algorithmic}
\end{algorithm}

\subsection{MoGE with Off-policy RL}
\subsubsection{Sample Quality of MoGE}
\label{sec: quality analyses}
As a generative exploration framework, MoGE synthesizes novel transitions that extend beyond the replay buffer. In what follows, we examine two key questions that underpin its design and effectiveness:

\emph{ (1) How does the generated transitions contribute to exploration and policy improvement?}

\emph{ (2) Will the generated transitions influence the policy improvement and evaluation of the algorithm?}

\paragraph{(a) Novelty and Policy-Dependent Generation.}
Let $\mathcal{D}_{\text{replay}}$ denote the replay buffer induced by a historical behavior policy $\beta$, corresponding to a state-action distribution $\rho_{\text{replay}}(s,a)$.
Conventional passive exploration frameworks~\cite{schaulPrioritized2016,lu2023synthetic,wang2024prioritized} intensify transitions from the replay buffer by reordering or imitating, remaining inherently constrained by $\beta$.
Consequently, the generated transitions $(s,a,s') \!\sim\! \rho$ cannot escape the coverage bias of the behavior policy.
In contrast, MoGE explicitly constructs a generative distribution
\begin{equation}
    \rho_{\text{gen}}(s,a) = p_{\varphi}(s)\,\pi_{\theta}(a|s),
\end{equation}
where $p_{\varphi}(s)$ denotes a learned state generator guided under a time-varying \emph{utility landscape} $f(s,\pi_{\theta}(s))$.
The critical state distribution is refined as follows:
\begin{equation}
 p_\varphi(s) \propto p_\varphi(s)\, \exp\!\big(\omega\, f(s,\pi_\theta(s))\big),
\end{equation}
Because $f$ evolves alongside the policy and critic, the target distribution for generation continuously shifts as the policy updates.
This property induces \textbf{continual novelty}---MoGE adaptively generates new, high-utility regions as the policy changes, rather than remaining tied to the historical behavior distribution.
Therefore, MoGE effectively decouples generation from the replay buffer, enabling exploration guided by both the \emph{current policy} and the \emph{estimated value function} instead of fully resampling or imitating past experiences.

\paragraph{(b) Dynamic Consistency and Bellman Validity.}
While novelty broadens exploration coverage, it must coexist with \emph{dynamical validity} to ensure that generated transitions remain consistent with the environment’s transition kernel $p_{\text{env}}(s'|s,a)$.
To this end, MoGE employs a learned world model $p_{\psi}(s'|s,a)$ enforcing the transition-level relation
\begin{equation}
    (s,a,s') \sim F_{\phi}(s'|s,a) \quad \text{such that} \quad s' \approx f_{\text{env}}(s,a),
\end{equation}
where $f_{\text{env}}$ denotes the true environment dynamics.
This mechanism ensures that the generated samples adhere to the underlying physical or causal transition relationship.

In contrast, methods that directly synthesize or interpolate transitions (e.g., via diffusion or latent interpolation) without explicit dynamics modeling can easily introduce spurious correlations among $(s,a,s')$ (loss functions that imitate the replay buffer do not ensure that valid transition dependencies are preserved among samples).
Such correlations violate the Markov property and lead to inconsistent TD estimates:
\begin{equation}
    \mathbb{E}_{(s,a,s')\sim\rho_{\text{syn}}}\!\left[ r(s,a) + \gamma V(s') \right]
    \;\neq\; \mathcal{T}^{\pi}V(s),
\end{equation}
where $\mathcal{T}^{\pi}$ is the Bellman operator.
These off-manifold transitions break the Bellman consistency and yield biased policy evaluation.
By leveraging the world model $F_{\phi}$, MoGE ensures that generated transitions satisfy the Bellman-consistent transition relation, thus preserving both physical and statistical validity.

\paragraph{(c) Summary.}
In summary, MoGE achieves \emph{continual novelty} via evolving utility-guided generation that adapts to the current policy, while maintaining \emph{dynamic consistency} through model-based transition regularization.
This synergy enables MoGE to explore beyond the behavior policy’s support without violating Bellman validity, leading to more reliable and effective policy learning.
\subsubsection{Training Off-policy Algorithm with MoGE}
\label{sec:mgpe}
To enhance exploration alongside policy improvement and evaluation by introducing the MoGE-generated samples in an off-policy manner, we propose a training framework that integrates MoGE into existing off-policy RL algorithms. Taking Actor-Critic as an example, the training framework is illustrated in Algorithm \ref{alg:mac}.

Notably, due to the existence of the critical generation, a distribution shift arises between the initial state distribution of the buffer and the generator. While this bias is negligible during policy evaluation, it must be addressed during policy improvement through Importance Sampling (IS). Since the distribution of the diffusion model cannot be explicitly represented, the importance sampling ratio for the initial state distribution is intractable. Therefore, we employ a sample mixing method to approximate importance sampling under bounded error. The formulas for policy evaluation and improvement are as follows:
\begin{equation}
\begin{aligned}
  &\mathcal L_{\text{PEV}}(\psi)=
  \mathbb E_{(s,a,r,s') \sim \mathcal D_k}
  \Bigl[
      Q_\psi(s,a) -
      \bigl(r + \gamma\,\mathbb E_{a' \sim \pi_\theta}
            [Q_{\bar \psi}(s',a')]\bigr)
  \Bigr]^2,\\
  &\mathcal L_{\text{PIM}}(\theta)
  =(1-\lambda)\mathbb{E}_{(s,a)\,\sim\,\mathcal D_{\text{env}}}
  \Bigl[
      g(s,a)
  \Bigr]
+
  \lambda\mathbb{E}_{(s,a)\,\sim\,\mathcal D_{\text{gen}}}
  \Bigl[
      g(s,a)
  \Bigr],
\end{aligned}
\end{equation}
where $D_k=(1-k)D_{\text{env}}+k D_{\text{gen}}$, $0 \leq k <1$, and $g(s, a)$ denotes the return in a single state-action pair $(s,a)$. It is worth noting that the error of this policy improvement method is controllable only when $\lambda$ is sufficiently small. The details of this approximation method, along with the selection of $k$ and $\lambda$, are referred to in the Appendix \ref{discussion 1}.

\section{Experiments}

\subsection{Experimental Setup.}
\textbf{Baselines.} We choose two widely used representative off-policy RL algorithms as our MoGE baselines: the stochastic policy algorithm DSAC \cite{duan2021distributional,duanDistributional2025}  and the deterministic policy algorithm TD3 \cite{fujimoto2018addressing}, which achieve active exploration through exploration bonus with policy entropy and random noise injection, respectively.
To further evaluate the performance of MoGE, we compare it with passive exploration methods like PGR \cite{wang2024prioritized} and prioritized experience replay (PER) \cite{schaulPrioritized2016}, where PGR is a state-of-the-art method in data augmentation that directly generates buffer samples using a diffusion model that has been shown to significantly enhance downstream learning.

\textbf{Benchmarks.} We evaluate our method on a diverse benchmark of 10 challenging locomotion tasks drawn from the DeepMind Control Suite (DMC) \cite{tassa2018deepmind} and the OpenAI Gym \cite{brockman2016openai}.  The Gym Benchmark introduces a wide range of control tasks. For example, the Humanoid environment raises the difficulty with high-dimensional state and action spaces (376/17 state/action dims). The DMC tasks feature complex agents: humanoid tasks (67/24 state/action dims) and quadruped tasks (78/12 state/action dims) that demand sophisticated balance and coordination.

\textbf{Implementation details.} To validate the plug-in capability of the MoGE, we preserve the original off-policy algorithms without further fine-tuning. In this paper, the total training step size for all experiments is set at 1.5 million, with the results of all experiments averaged over 3 random seeds. All hyperparameters are aligned with standard implementations, and the configuration details are documented in the Appendix \ref{app: details}.
\begin{figure}[t]
  \centering
  \begin{subfigure}[b]{0.195\textwidth}
    \includegraphics[width=\textwidth,trim={0.18cm 0.15cm 0.18cm 0.15cm},   clip]{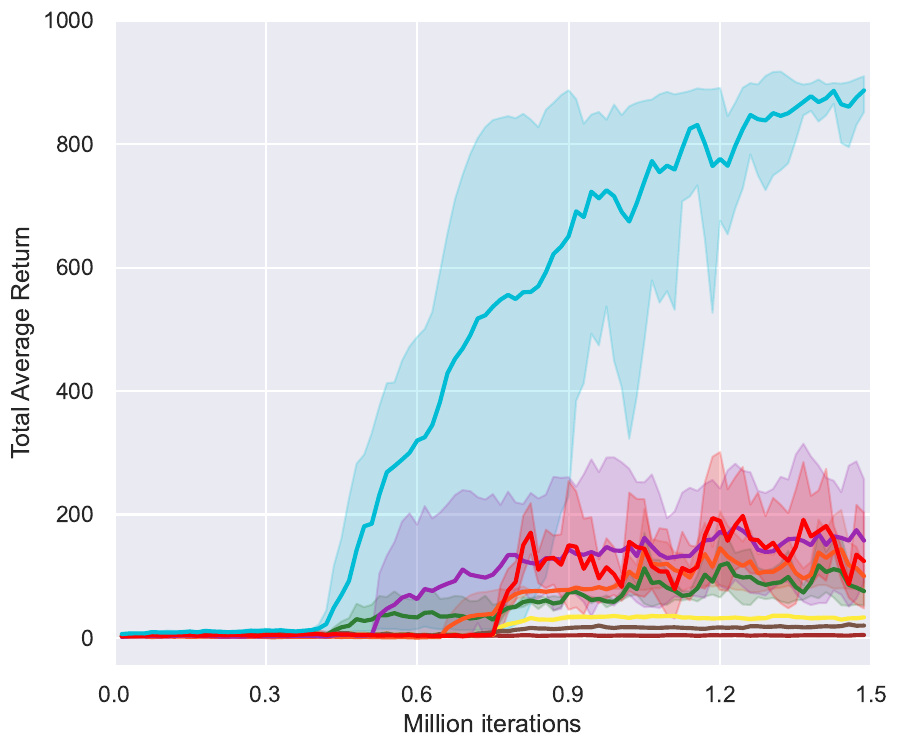}
    \caption{Humanoid-walk}
    \label{fig:humanoid_walk}
  \end{subfigure}
  \begin{subfigure}[b]{0.195\textwidth}
    \includegraphics[width=\textwidth,trim={0.18cm 0.15cm 0.18cm 0.15cm},   clip]{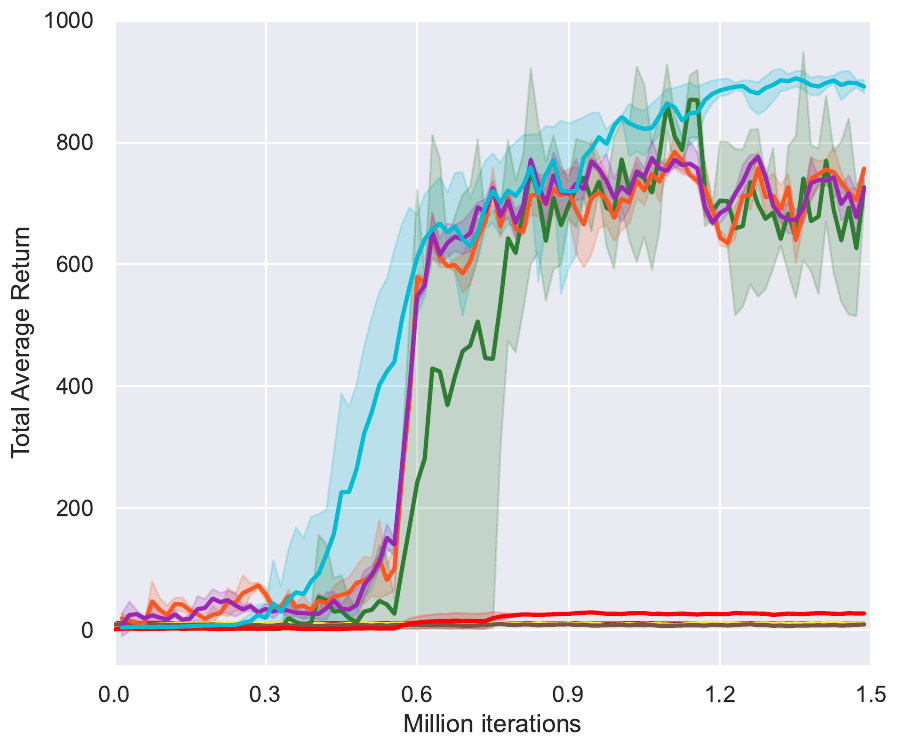}
    \caption{Humanoid-stand}
    \label{fig:humanoid-stand}
  \end{subfigure}
  \begin{subfigure}[b]{0.195\textwidth}
    \includegraphics[width=\textwidth,trim={0.18cm 0.15cm 0.18cm 0.15cm},   clip]{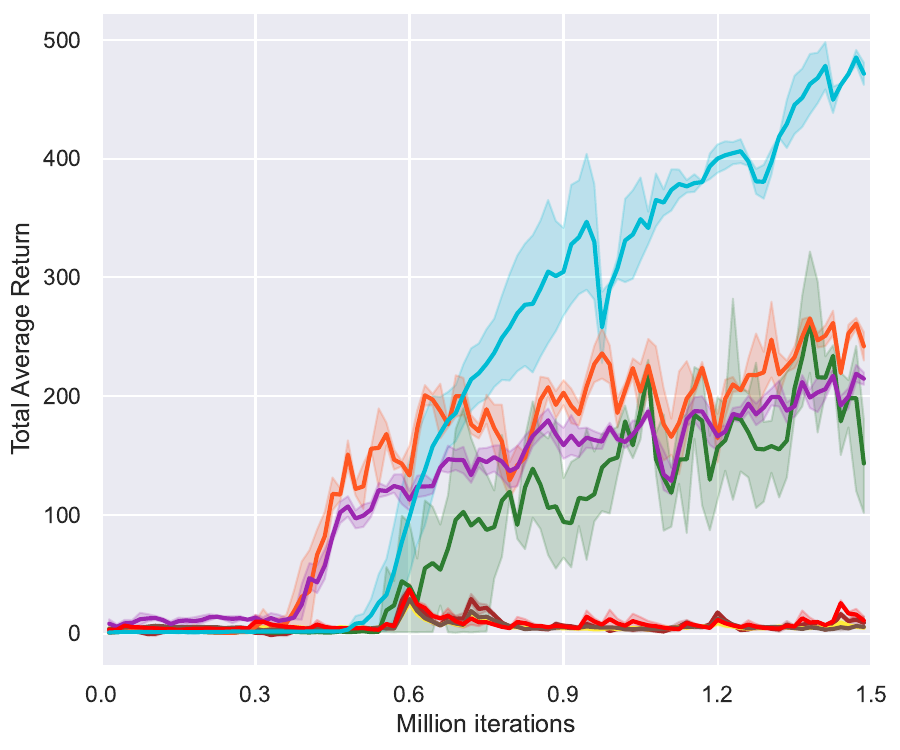}
    \caption{Humanoid-run}
    \label{fig:humanoid-run}
  \end{subfigure}
  \begin{subfigure}[b]{0.195\textwidth}
    \includegraphics[width=\textwidth,trim={0.18cm 0.15cm 0.18cm 0.15cm},   clip]{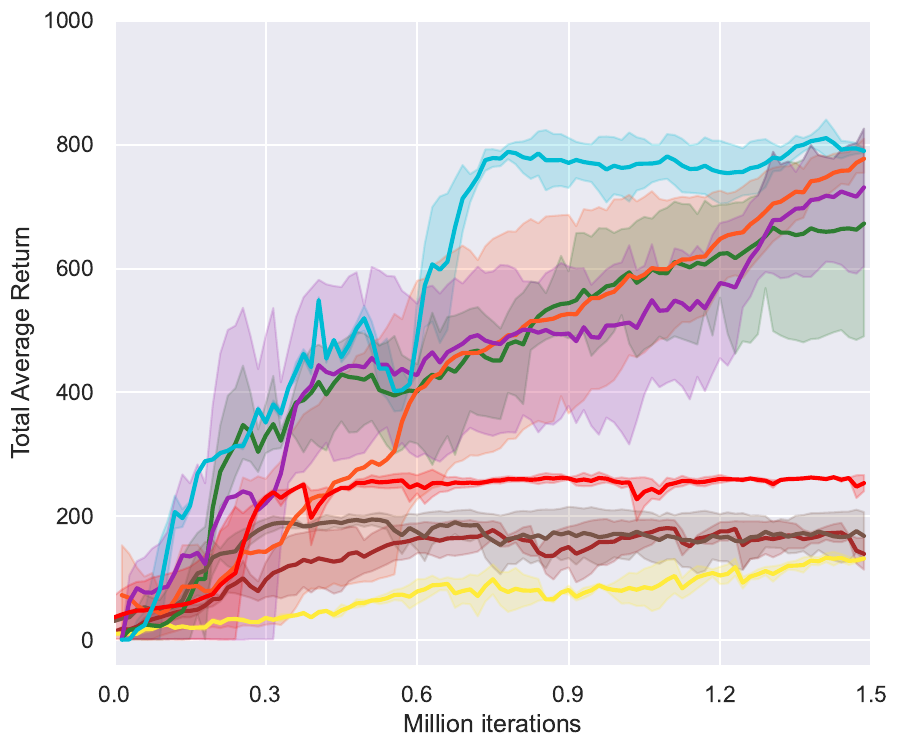}
    \caption{Quadruped-run}
    \label{Quadruped-run}
  \end{subfigure}
  \begin{subfigure}[b]{0.195\textwidth}
    \includegraphics[width=\textwidth,trim={0.18cm 0.15cm 0.18cm 0.15cm},   clip]{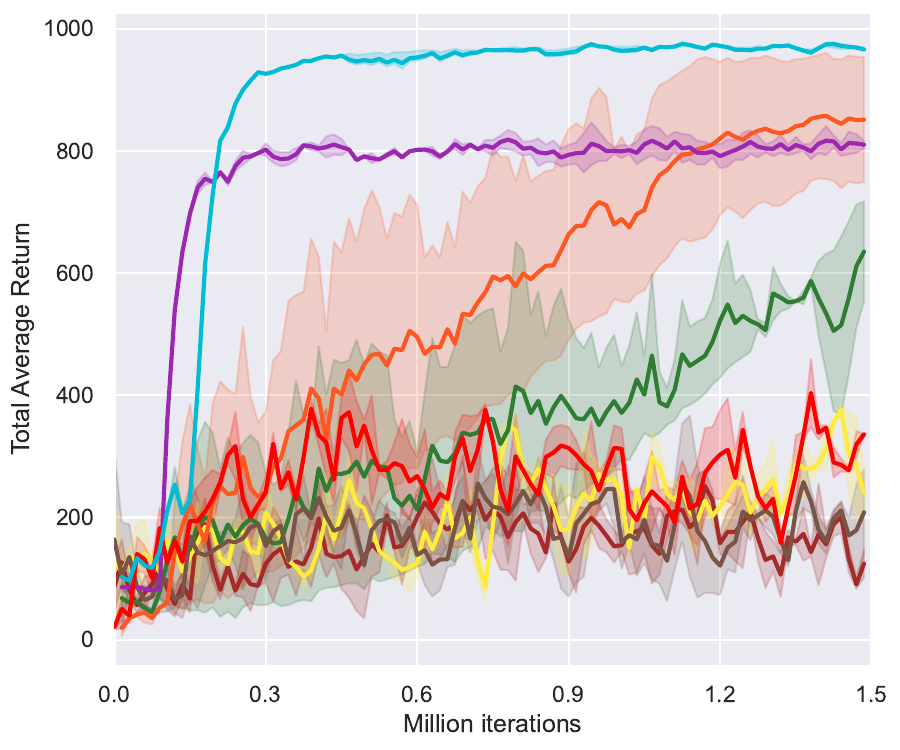}
    \caption{Quadruped-walk}
    \label{Quadruped-walk}
    \end{subfigure} \\
    \begin{subfigure}[b]{0.195\textwidth}
    \includegraphics[width=\textwidth,trim={0.18cm 0.15cm 0.18cm 0.15cm},   clip]{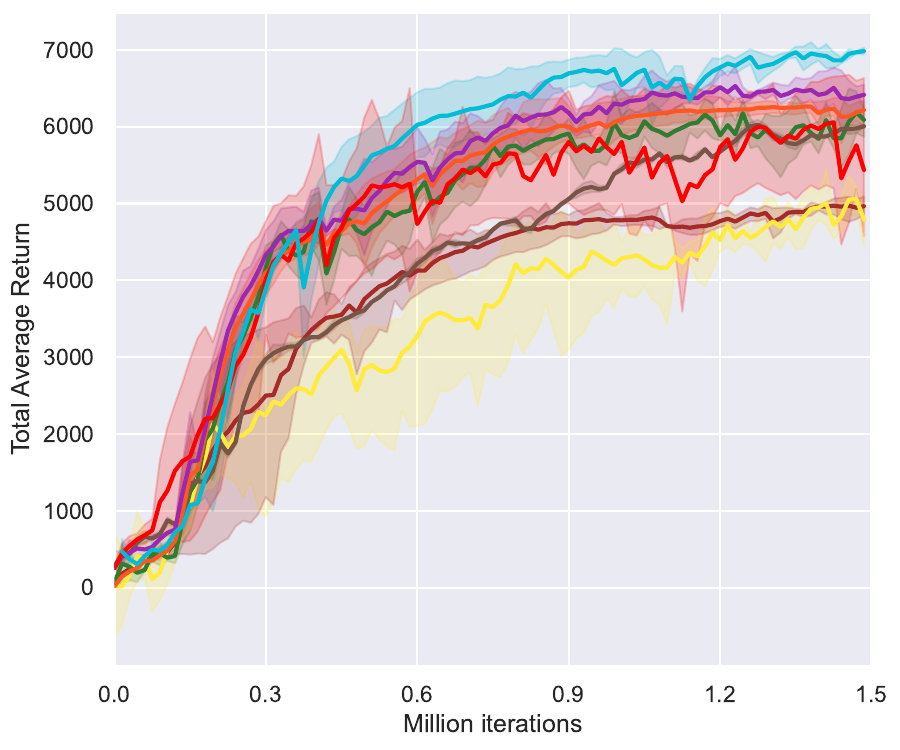}
    \caption{Walker2d-v3}
    \label{fig:walker2d-v3}
  \end{subfigure}
  \begin{subfigure}[b]{0.195\textwidth}
    \includegraphics[width=\textwidth,trim={0.18cm 0.15cm 0.18cm 0.15cm},   clip]{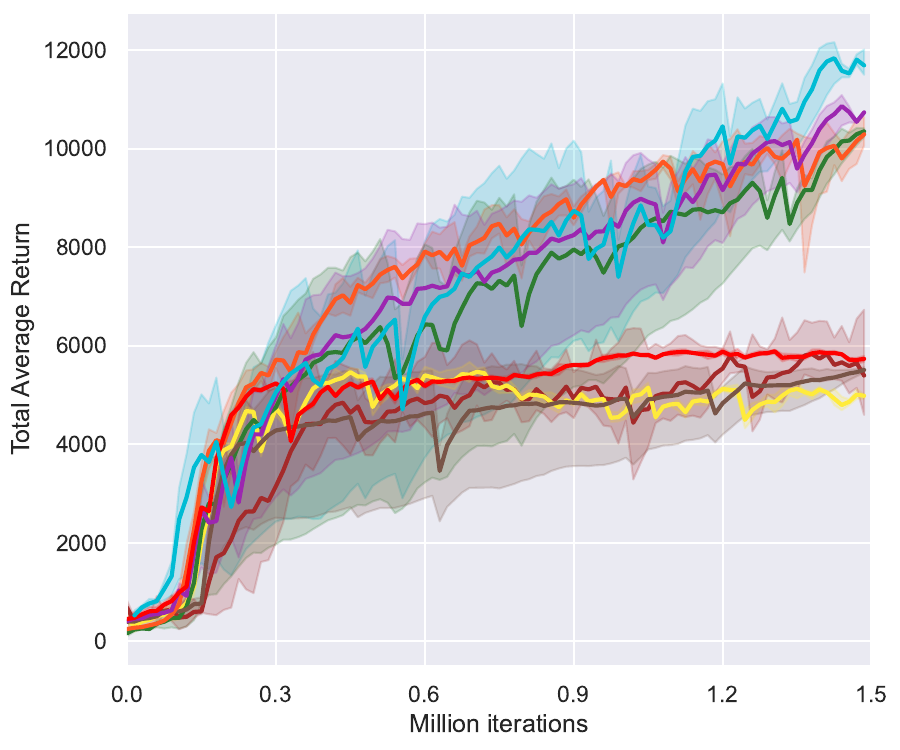}
    \caption{Humanoid-v3}
    \label{fig:humanoid-v3}
  \end{subfigure}
  \begin{subfigure}[b]{0.195\textwidth}
    \includegraphics[width=\textwidth,trim={0.18cm 0.15cm 0.18cm 0.15cm},   clip]{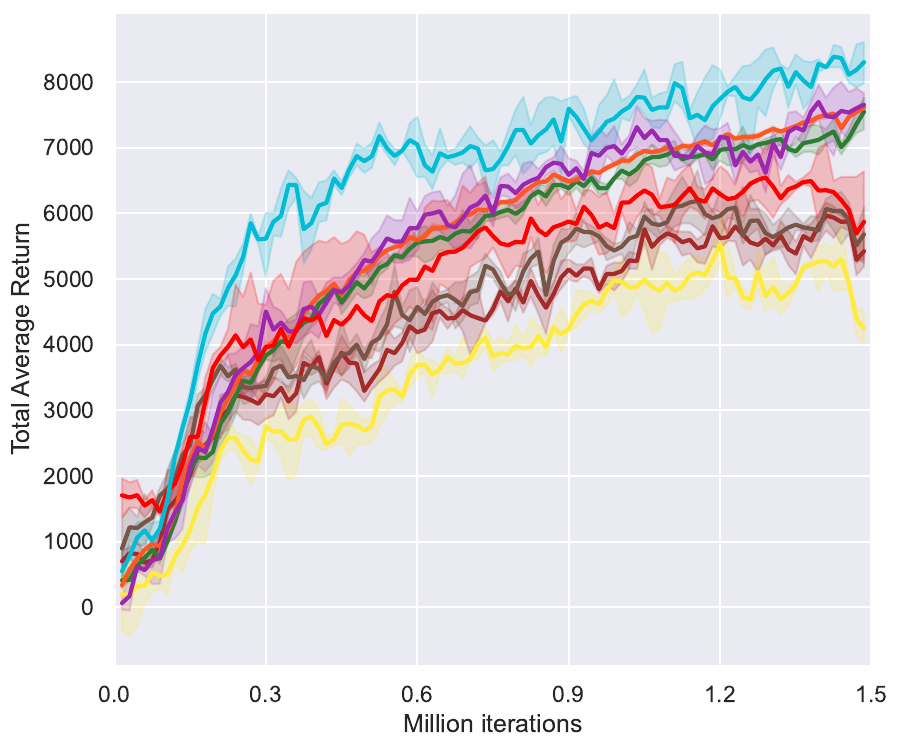}
    \caption{Ant-v3}
    \label{fig:ant-v3}
  \end{subfigure}
  \begin{subfigure}[b]{0.195\textwidth}
    \includegraphics[width=\textwidth,trim={0.18cm 0.15cm 0.18cm 0.15cm},   clip]{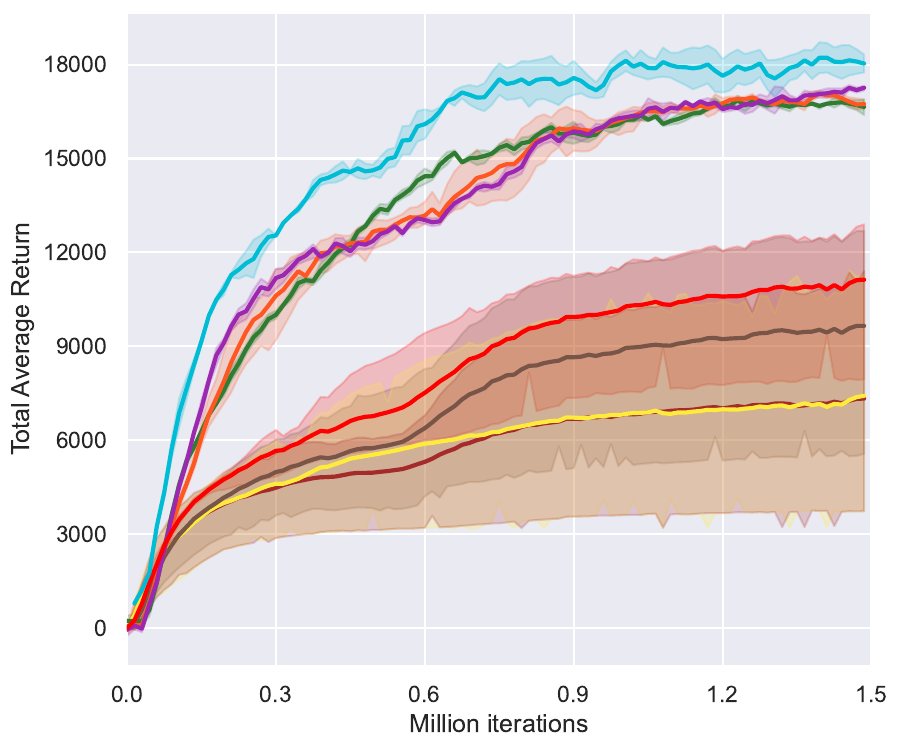}
    \caption{Halfcheetah-v3}
    \label{fig:halfcheetah-v3}
  \end{subfigure}
  \begin{subfigure}[b]{0.195\textwidth}
    \includegraphics[width=\textwidth,trim={0.18cm 0.15cm 0.18cm 0.15cm},   clip]{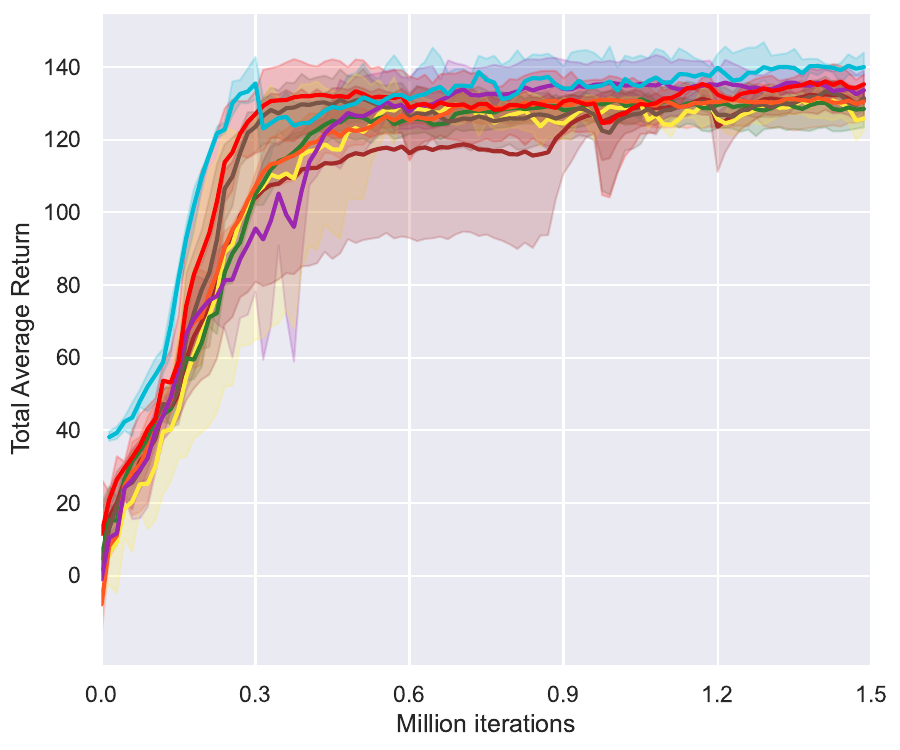}
    \caption{Swimmer-v3}
    \label{fig:Swimmer-v3}
  \end{subfigure}\\
  \begin{subfigure}[b]{\textwidth}
    \centering
    \includegraphics[width=0.9\textwidth,trim={0.1cm 0.1cm 0.3cm 0.1cm},   clip]{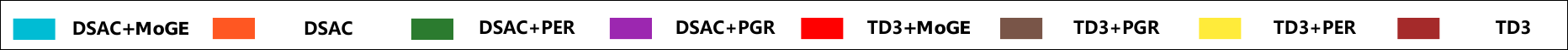}
  \end{subfigure}
  \caption{\textbf{Training curves on benchmarks.} The solid lines depict the mean performance, while the shaded areas represent the confidence intervals over three seeds. The first row corresponds to the training curves on the DeepMind Control Suite, while the second row represents the results on OpenAI Gym.}
  \label{fig:training_curves}
\end{figure}

\subsection{Experimental Results}
All the training curves are shown in Figure \ref{fig:training_curves} and the detailed results are listed in Table~\ref{tab:comparison}.
Our method, \textbf{MoGE}, consistently achieves superior Total Average Return (TAR) across a wide range of locomotion tasks. Despite the challenges introduced by high-dimensional state and action spaces as well as intricate dynamics, it maintains exceptional stability and efficiency, highlighting its robustness and adaptability.

In the challenging DMC Suite tasks, MoGE demonstrates substantial performance enhancements over the original TD3 and DSAC algorithms.  MoGE achieves an average Total Average Return (TAR) of \textbf{817.7}, significantly outperforming the original DSAC method (568.5) by a notable \textbf{+43.8\%}. In individual tasks, such as \textit{Humanoid-walk}, MoGE reaches \textbf{891.7}, a remarkable \textbf{+508.6\%} improvement over the original DSAC (146.5). Similarly,  TD3 with MoGE significantly surpasses original TD3, delivering a \textbf{+73.3\%} improvement.

In the OpenAI Gym tasks, MoGE continues to exhibit exceptional performance.  MoGE achieves an average TAR of \textbf{9135.5}, surpassing the standard DSAC (8301.0) by \textbf{+10.0\%}. Notably, MoGE sets new benchmark results across all evaluated environments. In \textit{Humanoid-v3}, it attains a score of \textbf{12151.1}, a substantial \textbf{+16.8\%} increase over DSAC (10402.2). MoGE's consistent performance highlights its superior effectiveness, offering clear and significant improvements in both early-stage learning and final asymptotic returns compared to traditional TD3 and DSAC implementations.

\begin{table}[h]
\centering
\fontsize{6pt}{10}\selectfont
\setlength{\tabcolsep}{2pt}
\caption{Total Average Return (TAR) on 5 DMC Suite tasks and 5 OpenAI Gym tasks. Mean ± Std over 3 seeds. \textbf{Bold} = best; Higher is better.}
\begin{tabular}{lcccccccccc}
\toprule
\textbf{Environment}
& TD3 & TD3+PER & TD3+PGR & TD3+MoGE & DSAC & DSAC+PER & DSAC+PGR & DSAC+MoGE \\
\midrule
Humanoid-walk & 120.1 ± 106.8 & 34.4 ± 1.2 & 22.5 ± 2.4 & 222.4 ± 72.0 & 146.5 ± 60.9 & 122.5 ± 48.5 & 195.0 ± 92.3 & \textbf{891.7 ± 19.1} \\
Humanoid-stand    & 10.6 ± 0.2 & 11.5 ± 0.9 & 8.3 ± 0.2 & 28.4 ± 0.7 & 776.6 ± 15.6 & 816.5 ± 94.5 & 754.4 ± 16.2 & \textbf{907.5 ± 6.9} \\
Humanoid-run        & 15.9 ± 1.8 & 8.7 ± 2.8 & 7.0 ± 0.4 & 25.2 ± 2.5 & 267.4 ± 3.9 & 271.1 ± 35.5 & 223.4 ± 3.3 & \textbf{488.9 ± 8.7} \\
Quadruped-run      & 84.1 ± 5.4 & 63.3 ± 2.5 & 83.2 ± 16.6 & 128.2 ± 0.4 & 793.9 ± 29.9 & 662.7 ± 162.6 & 717.9 ± 107.4 & \textbf{824.3 ± 19.0} \\
Quadruped-walk  &236.5 ± 19.5 & 380.2 ± 7.2 & 290.2 ± 11.7 & 405.2 ± 54.7 & 857.9 ± 102.8 & 649.3 ± 67.7 & 823.2 ± 17.5 & \textbf{976.2 ± 3.1} \\
\midrule
AVG.DMC      & 93.4 ± 26.7 & 99.62 ± 2.9 & 82.2 ± 6.3& 161.9 ± 26.1 & 568.5±42.7 & 504.4 ± 81.8 & 542.8 ± 47.3 & \textbf{817.7 ± 11.4 } \\
\midrule
Walker2d-v3  & 5031.1 ± 84.2 & 5253.7 ± 206.1 & 6007.5 ± 4.9 & 6082.8 ± 606.7 & 6288.3 ± 83.3 & 6391.6 ± 246.9 & 6501.1 ± 87.3 & \textbf{6978.4 ± 68.7 }\\
Humanoid-v3    & 5967.1 ± 547.8 & 5203.9 ± 33.9 & 5531.8 ± 62.3 & 5885.8 ± 38.3 & 10402.2 ± 187.7 & 10363.6 ± 109.0 & 11004.0 ± 121.5 & \textbf{12151.1 ± 35.4} \\
Ant-v3       & 6037.5 ± 119.2 & 6134.7 ± 212.5 & 5961.3 ± 170.7 & 6369.1 ± 265.7 & 7610.0 ± 10.0 & 7637.0 ± 27.1 & 7837.2 ± 203.0 & \textbf{8176.6 ± 44.9} \\
Halfcheetah-v3    &  7363.0 ± 3666.4 & 7438.1 ± 3774.3 & 9687.1 ± 4099.0 & 11167.3 ± 2234.8 & 17072.0 ± 61.3 & 16913.0 ± 70.2 & 17324.7 ± 41.1 & \textbf{18054.9 ± 459.6} \\
Swimmer-v3    &   133.5 ± 5.3 & 131.8 ± 3.3 & 134.1 ± 2.8 & 137.1 ± 2.8 & 132.3 ± 5.1 & 131.0 ± 6.6 & 136.0 ± 3.0 & \textbf{141.3 ± 2.0} \\
\midrule
AVG.Gym      & 4906.4 ± 884.6 & 4832.4 ± 846.0 & 5464.4 ± 867.9 & 5928.4 ± 629.7 & 8301.0 ± 69.5 & 8287.2 ± 92.0 & 8535.6 ± 91.2 & \textbf{9135.5 ± 122.1} \\
\bottomrule
\end{tabular}
\label{tab:comparison}
\end{table}

\subsection{Ablation}
\begin{figure}[h]
  \centering
  \begin{subfigure}[b]{0.3\textwidth}
    \includegraphics[width=\textwidth,trim={0.18cm 0.15cm 0.18cm 0.15cm},   clip]{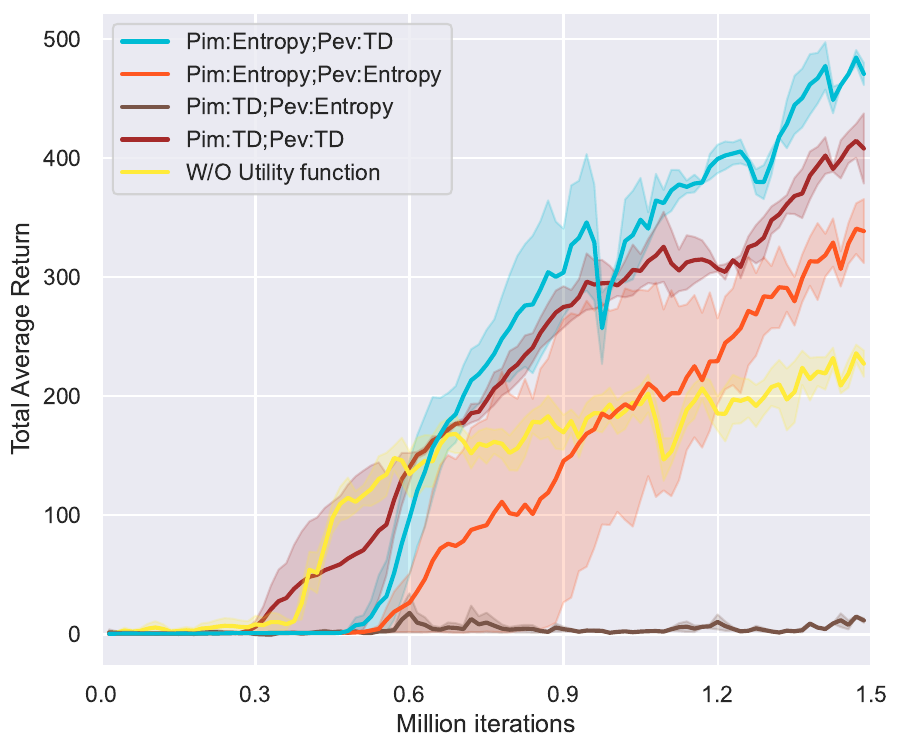}
    \captionsetup{width=0.75\linewidth}
    \caption{Ablation study on the choice of utility function}
    \label{fig:Ablation1}
  \end{subfigure}
  \begin{subfigure}[b]{0.3\textwidth}
    \includegraphics[width=\textwidth,trim={0.18cm 0.15cm 0.18cm 0.15cm},   clip]{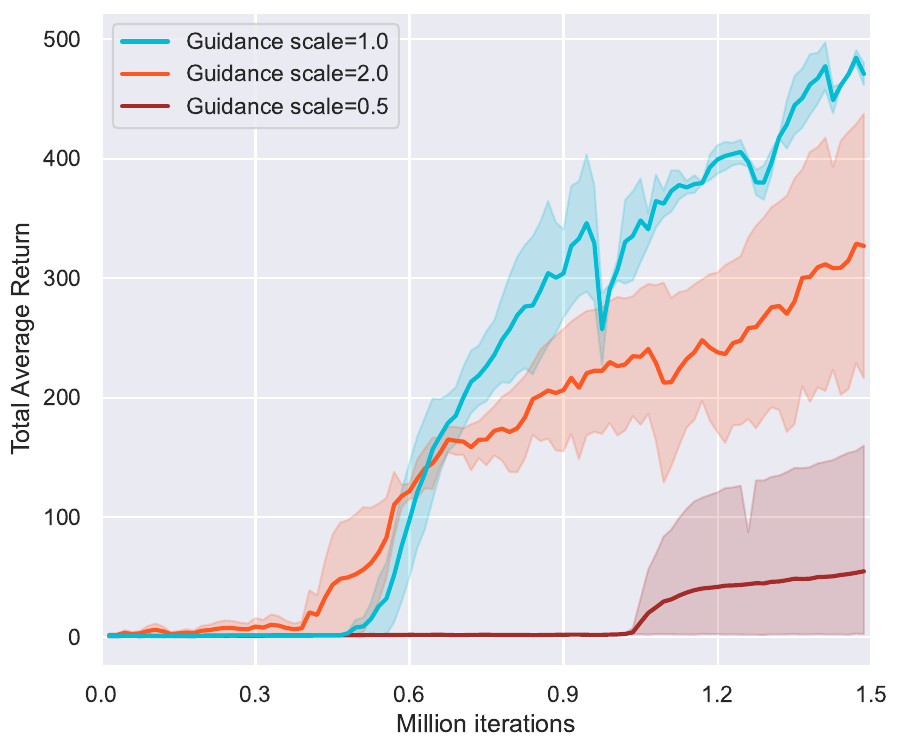}
    \captionsetup{width=0.75\linewidth}
    \caption{Ablation study on the guidance scale}
    \label{fig:Ablation2}
  \end{subfigure}
  \begin{subfigure}[b]{0.3\textwidth}
    \includegraphics[width=\textwidth,trim={0.18cm 0.15cm 0.18cm 0.15cm},   clip]{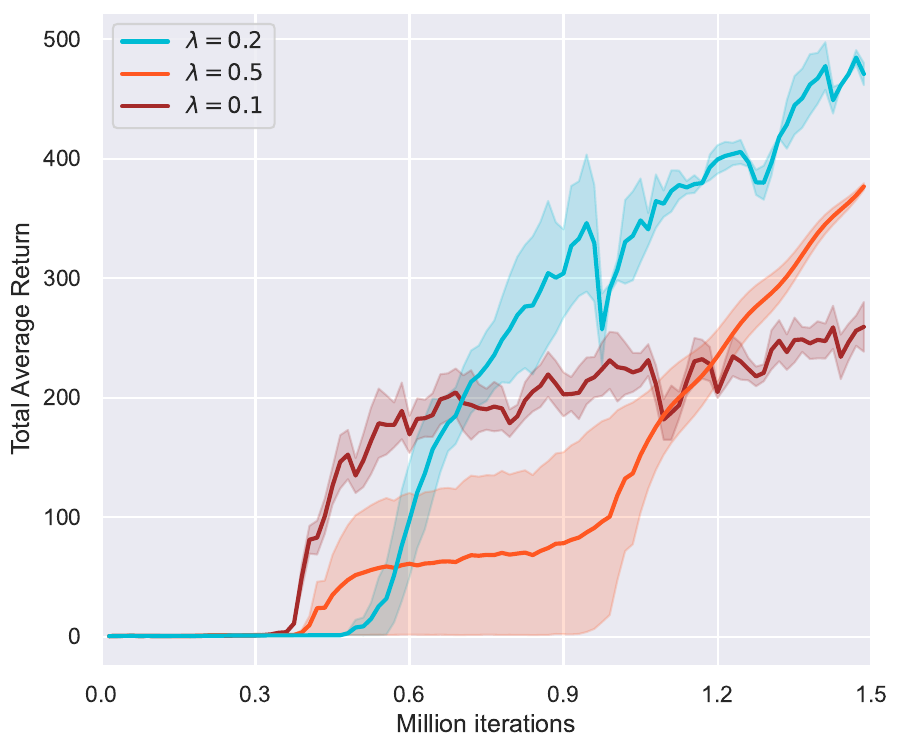}
    \captionsetup{width=0.75\linewidth}
    \caption{Ablation study on the mix ratio}
    \label{fig:Ablation3}
  \end{subfigure}

  \caption{\textbf{Ablation study curves.} We select the \textit{Humanoid-run} task in DMC Suite with high complexity to perform all ablation experiments.}
  \label{fig:example_3_images}
\end{figure}

We perform three ablation studies to evaluate the impact of each core component in our framework:

\textbf{Utility function for exploration.} \
We compare the choice of utility function for different parts of updating, as illustrated in Figure \ref{fig:Ablation1}. Compared to policy entropy, TD error is more beneficial during policy evaluation. On the other hand, entropy plays a more significant role in policy improvement since high-entropy regions encourage broader exploration, while high TD-error regions may stem from inaccurate environment estimation, potentially leading to unreliable evaluations.

\textbf{Guidance scale $\omega$.} \
We chose different guidance intensities to test the balance between the regions with high potential for exploration and feasibility guarantee, as shown in Figure~\ref{fig:Ablation2}. When the guidance strength is set to 1, it effectively balances the generation of states with high exploratory value and feasibility, ensuring that the diffusion-generated states maintain alignment with the optimal occupancy measure while maximizing exploration potential.

\textbf{Mix ratio $\lambda$.} \
We vary the value of $\lambda$ by testing from $0.1$ to $0.5$. Results in Figure~\ref{fig:Ablation3} show that $\lambda=0.2$ stable performance across this range. When $\lambda$ is too small, the policy fails to acquire enough critical states for effective exploration. Conversely, if $\lambda$ is too large, the discrepancy in the state distribution becomes non-negligible, which aligns with the discussion in Appendix \ref{discussion 1}.

\section{Related Works}

\textbf{Active exploration.}
Active exploration methods explicitly modify policies to enhance exploration, which originates from the epsilon-greedy policy \cite{sutton1998reinforcement, mnihHumanlevel2015, plappert2017parameter, fujimoto2018addressing}. Entropy-based strategies like SAC \cite{haarnojaSoft2018}, DSAC \cite{duanDistributional2025}, DACER \cite{wang2025enhanced, wang2024diffusion}, and PPO \cite{schulmanProximal2017} incorporate entropy terms to balance exploration and exploitation, preventing premature convergence and enhancing stability.
More advanced exploration approaches explicitly encourage policy diversity. Count-based methods, such as pseudo-counts \cite{bellemareUnifying2016} and neural density models \cite{ostrovskiCountBased2017}, estimate state visitation frequencies to incentivize exploration in under-sampled regions. Intrinsic motivation strategies, including RND \cite{burdaExploration2019} and ICM \cite{pathakCuriosityDriven2017}, generate exploration bonuses from novelty signals or prediction errors. Bootstrap DQN \cite{osbandDeep2016} further introduces uncertainty-aware exploration by leveraging ensemble networks to identify poorly understood state-action pairs.
Recent advances maintain exploration diversity through maximum entropy principles, and exploration-driven policy optimization \cite{abdolmalekiMaximum2018,pathakCuriosityDriven2017}, which directly augments policies for broader state coverage. These methods, while easy to implement, often require complex parameter tuning and may struggle with scalability in high-dimensional spaces compared with MoGE.

\textbf{Passive exploration.}
The policy update process not only depends on its optimization objective, but also on the samples collected for estimating the gradient. By shifting the distribution of replayed samples, the policy is able to cover a wider range of states. Prioritized experience replay (PER) \cite{schaulPrioritized2016} dynamically adjusts the replaying frequency of collected samples by TD-error. Some further research \cite{ramicicEntropybased2017, sujitPrioritizing2023} uses different metrics for experience replay. Rather than only selecting the samples, data generation is more flexible since it not only models the initial state but also captures its transition. Diffuser \cite{jannerPlanning2022} utilizes a diffusion probabilistic model that plans by iteratively denoising trajectories. Synthetic experience replay (SER) \cite{lu2023synthetic} proposes a diffusion-based approach to flexibly upsample an agent’s collected experience. Prioritized generative replay (PGR) \cite{wang2024prioritized} generates learning-informative transitions under a given relevance function. Another paradigm of passive exploration is model-based RL \cite{ha2018world}. These methods learns a parameterized transition model of state and action and directly optimize the policy through imagined samples, such as Dreamer \cite{hafner2019dream, hafner2020mastering, hafner2023mastering} and other variants \cite{chen2022transdreamer, hansenTemporal2022, zhangSTORM2023, hansenTDMPC22024}. Our proposed MoGE method can generate critical initial states and their transitions, thus guaranteeing data compliance.

\section{Conclusion}

We proposed MoGE, a novel exploration paradigm that addresses the limitations of passive exploration in off-policy RL. It enhances exploration by generating critical states guided by exploratory priors and then estimates state transitions through a world model, forming valid training samples that guarantee state-space compliance and dynamical feasibility. Experimental results on OpenAI Gym and DeepMind Control Suite benchmarks demonstrate that MoGE significantly improves sample efficiency and overall performance, validating the effectiveness of this exploration paradigm.
We believe MoGE establishes a new perspective for exploration augmentation in reinforcement learning, with significant potential for future improvements, such as incorporating adaptive mechanisms for prioritizing critical state transitions or leveraging more expressive generative models beyond conditional diffusion to further boost exploration efficiency and policy robustness.

\section{Acknowledgment}
This research was generously supported by the Beijing Natural Science Foundation (L257002). The core of this work was completed in the Intelligence Driving Lab (IDLAB) at Tsinghua University. We extend our sincere gratitude to all the lab members for their constructive feedback during this period.

\newpage
\bibliographystyle{plain}
\bibliography{neurips_2025}

\begin{thebibliography}{10}

\bibitem{abdolmalekiMaximum2018}
Abbas Abdolmaleki, Yunlong Huang, Timothy Lau, Leonard Hasenclever, Michael Neunert, Ioannis Tsiamas, Abbas Abdolmaleki, Marc Deisenroth, Martin Riedmiller, and Nicolas Heess.
\newblock Maximum a posteriori policy optimisation.
\newblock In {\em Proceedings of the International Conference on Learning Representations (ICLR)}, 2018.

\bibitem{abdolmaleki2018max}
Abbas Abdolmaleki, Jost~Tobias Springenberg, Yuval Tassa, Remi Munos, Nicolas Heess, and Martin Riedmiller.
\newblock Maximum a posteriori policy optimisation, 2018.

\bibitem{ahmed2019understanding}
Zafarali Ahmed, Nicolas Le~Roux, Mohammad Norouzi, and Dale Schuurmans.
\newblock Understanding the impact of entropy on policy optimization.
\newblock In {\em International conference on machine learning}, pages 151--160. PMLR, 2019.

\bibitem{banerjee2023boosting}
Chayan Banerjee, Zhiyong Chen, and Nasimul Noman.
\newblock Boosting exploration in actor-critic algorithms by incentivizing plausible novel states.
\newblock In {\em 2023 62nd IEEE Conference on Decision and Control (CDC)}, pages 7009--7014. IEEE, 2023.

\bibitem{bellemareUnifying2016}
Marc~G. Bellemare, Sriram Srinivasan, Georg Ostrovski, Tom Schaul, David Saxton, and Remi Munos.
\newblock Unifying count-based exploration and intrinsic motivation.
\newblock In {\em Proceedings of the International Conference on Machine Learning (ICML)}, 2016.

\bibitem{brockman2016openai}
Greg Brockman, Vicki Cheung, Ludwig Pettersson, Jonas Schneider, John Schulman, Jie Tang, and Wojciech Zaremba.
\newblock Openai gym.
\newblock {\em arXiv preprint arXiv:1606.01540}, 2016.

\bibitem{burdaExploration2019}
Yuri Burda, Harrison Edwards, Amos Storkey, and Oleg Klimov.
\newblock Exploration by random network distillation.
\newblock In {\em Proceedings of the International Conference on Learning Representations (ICLR)}, 2019.

\bibitem{chen2022transdreamer}
Chang Chen, Yi-Fu Wu, Jaesik Yoon, and Sungjin Ahn.
\newblock Transdreamer: Reinforcement learning with transformer world models.
\newblock {\em arXiv preprint arXiv:2202.09481}, 2022.

\bibitem{chen2022redeeming}
Eric Chen, Zhang-Wei Hong, Joni Pajarinen, and Pulkit Agrawal.
\newblock Redeeming intrinsic rewards via constrained optimization.
\newblock {\em Advances in neural information processing systems}, 35:4996--5008, 2022.

\bibitem{chen2021interpretable}
Jianyu Chen, Shengbo~Eben Li, and Masayoshi Tomizuka.
\newblock Interpretable end-to-end urban autonomous driving with latent deep reinforcement learning.
\newblock {\em IEEE Transactions on Intelligent Transportation Systems}, 23(6):5068--5078, 2021.

\bibitem{chen2021randomized}
Xinyue Chen, Che Wang, Zijian Zhou, and Keith Ross.
\newblock Randomized ensembled double q-learning: Learning fast without a model.
\newblock {\em arXiv preprint arXiv:2101.05982}, 2021.

\bibitem{dhariwal2021diffusion}
Prafulla Dhariwal and Alexander Nichol.
\newblock Diffusion models beat gans on image synthesis.
\newblock {\em Advances in neural information processing systems}, 34:8780--8794, 2021.

\bibitem{dogru2024reinforcement}
Oguzhan Dogru, Junyao Xie, Om~Prakash, Ranjith Chiplunkar, Jansen Soesanto, Hongtian Chen, Kirubakaran Velswamy, Fadi Ibrahim, and Biao Huang.
\newblock Reinforcement learning in process industries: Review and perspective.
\newblock {\em IEEE/CAA Journal of Automatica Sinica}, 11(2):283--300, 2024.

\bibitem{duan2021distributional}
Jingliang Duan, Yang Guan, Shengbo~Eben Li, Yangang Ren, Qi~Sun, and Bo~Cheng.
\newblock Distributional soft actor-critic: Off-policy reinforcement learning for addressing value estimation errors.
\newblock {\em IEEE transactions on neural networks and learning systems}, 33(11):6584--6598, 2021.

\bibitem{duanDistributional2025}
Jingliang Duan, Wenxuan Wang, Liming Xiao, Jiaxin Gao, Shengbo~Eben Li, Chang Liu, Ya-Qin Zhang, Bo~Cheng, and Keqiang Li.
\newblock Distributional {{Soft Actor-Critic With Three Refinements}}.
\newblock 47(5):3935--3946.

\bibitem{eysenbach2019if}
Benjamin Eysenbach and Sergey Levine.
\newblock If maxent rl is the answer, what is the question?
\newblock {\em arXiv preprint arXiv:1910.01913}, 2019.

\bibitem{flennerhag2020temporal}
Sebastian Flennerhag, Jane~X Wang, Pablo Sprechmann, Francesco Visin, Alexandre Galashov, Steven Kapturowski, Diana~L Borsa, Nicolas Heess, Andre Barreto, and Razvan Pascanu.
\newblock Temporal difference uncertainties as a signal for exploration.
\newblock {\em arXiv preprint arXiv:2010.02255}, 2020.

\bibitem{fujimoto2018addressing}
Scott Fujimoto, Herke Hoof, and David Meger.
\newblock Addressing function approximation error in actor-critic methods.
\newblock In {\em International conference on machine learning}, pages 1587--1596. PMLR, 2018.

\bibitem{geng2025mean}
Zhengyang Geng, Mingyang Deng, Xingjian Bai, J~Zico Kolter, and Kaiming He.
\newblock Mean flows for one-step generative modeling.
\newblock {\em arXiv preprint arXiv:2505.13447}, 2025.

\bibitem{grondman2012survey}
Ivo Grondman, Lucian Busoniu, Gabriel~AD Lopes, and Robert Babuska.
\newblock A survey of actor-critic reinforcement learning: Standard and natural policy gradients.
\newblock {\em IEEE Transactions on Systems, Man, and Cybernetics, part C (applications and reviews)}, 42(6):1291--1307, 2012.

\bibitem{gupta2021embodied}
Agrim Gupta, Silvio Savarese, Surya Ganguli, and Li~Fei-Fei.
\newblock Embodied intelligence via learning and evolution.
\newblock {\em Nature communications}, 12(1):5721, 2021.

\bibitem{ha2018world}
David Ha and J{\"u}rgen Schmidhuber.
\newblock World models.
\newblock {\em arXiv preprint arXiv:1803.10122}, 2018.

\bibitem{haarnoja2018soft}
Tuomas Haarnoja, Aurick Zhou, Pieter Abbeel, and Sergey Levine.
\newblock Soft actor-critic: Off-policy maximum entropy deep reinforcement learning with a stochastic actor.
\newblock In {\em International conference on machine learning}, pages 1861--1870. Pmlr, 2018.

\bibitem{haarnojaSoft2018}
Tuomas Haarnoja, Aurick Zhou, Pieter Abbeel, and Sergey Levine.
\newblock Soft actor-critic: Off-policy maximum entropy deep reinforcement learning with a stochastic actor.
\newblock In Jennifer~G. Dy and Andreas Krause, editors, {\em Proceedings of the 35th International Conference on Machine Learning, {ICML} 2018, Stockholmsm{\"{a}}ssan, Stockholm, Sweden, July 10-15, 2018}, volume~80 of {\em Proceedings of Machine Learning Research}, pages 1856--1865. {PMLR}, 2018.

\bibitem{hafner2019dream}
Danijar Hafner, Timothy Lillicrap, Jimmy Ba, and Mohammad Norouzi.
\newblock Dream to control: Learning behaviors by latent imagination.
\newblock {\em arXiv preprint arXiv:1912.01603}, 2019.

\bibitem{hafner2020mastering}
Danijar Hafner, Timothy Lillicrap, Mohammad Norouzi, and Jimmy Ba.
\newblock Mastering atari with discrete world models.
\newblock {\em arXiv preprint arXiv:2010.02193}, 2020.

\bibitem{hafner2023mastering}
Danijar Hafner, Jurgis Pasukonis, Jimmy Ba, and Timothy Lillicrap.
\newblock Mastering diverse domains through world models.
\newblock {\em arXiv preprint arXiv:2301.04104}, 2023.

\bibitem{han2021diversity}
Seungyul Han and Youngchul Sung.
\newblock Diversity actor-critic: Sample-aware entropy regularization for sample-efficient exploration.
\newblock In {\em International Conference on Machine Learning}, pages 4018--4029. PMLR, 2021.

\bibitem{hansenTemporal2022}
Nicklas Hansen, Hao Su, and Xiaolong Wang.
\newblock Temporal difference learning for model predictive control.
\newblock In Kamalika Chaudhuri, Stefanie Jegelka, Le~Song, Csaba Szepesv{\'{a}}ri, Gang Niu, and Sivan Sabato, editors, {\em International Conference on Machine Learning, {ICML} 2022, 17-23 July 2022, Baltimore, Maryland, {USA}}, volume 162 of {\em Proceedings of Machine Learning Research}, pages 8387--8406. {PMLR}, 2022.

\bibitem{hansenTDMPC22024}
Nicklas Hansen, Hao Su, and Xiaolong Wang.
\newblock {TD-MPC2:} scalable, robust world models for continuous control.
\newblock In {\em The Twelfth International Conference on Learning Representations, {ICLR} 2024, Vienna, Austria, May 7-11, 2024}. OpenReview.net, 2024.

\bibitem{hiraoka2021dropout}
Takuya Hiraoka, Takahisa Imagawa, Taisei Hashimoto, Takashi Onishi, and Yoshimasa Tsuruoka.
\newblock Dropout q-functions for doubly efficient reinforcement learning.
\newblock {\em arXiv preprint arXiv:2110.02034}, 2021.

\bibitem{ho2020denoising}
Jonathan Ho, Ajay Jain, and Pieter Abbeel.
\newblock Denoising diffusion probabilistic models.
\newblock {\em Advances in neural information processing systems}, 33:6840--6851, 2020.

\bibitem{jannerPlanning2022}
Michael Janner, Yilun Du, Joshua~B. Tenenbaum, and Sergey Levine.
\newblock Planning with diffusion for flexible behavior synthesis.
\newblock In Kamalika Chaudhuri, Stefanie Jegelka, Le~Song, Csaba Szepesv{\'{a}}ri, Gang Niu, and Sivan Sabato, editors, {\em International Conference on Machine Learning, {ICML} 2022, 17-23 July 2022, Baltimore, Maryland, {USA}}, volume 162 of {\em Proceedings of Machine Learning Research}, pages 9902--9915. {PMLR}, 2022.

\bibitem{janz2019successor}
David Janz, Jiri Hron, Przemys{\l}aw Mazur, Katja Hofmann, Jos{\'e}~Miguel Hern{\'a}ndez-Lobato, and Sebastian Tschiatschek.
\newblock Successor uncertainties: exploration and uncertainty in temporal difference learning.
\newblock {\em Advances in Neural Information Processing Systems}, 32, 2019.

\bibitem{laroche2023occupancy}
Romain Laroche and Remi~Tachet Des~Combes.
\newblock On the occupancy measure of non-markovian policies in continuous mdps.
\newblock In {\em International Conference on Machine Learning}, pages 18548--18562. PMLR, 2023.

\bibitem{lee2024simba}
Hojoon Lee, Dongyoon Hwang, Donghu Kim, Hyunseung Kim, Jun~Jet Tai, Kaushik Subramanian, Peter~R Wurman, Jaegul Choo, Peter Stone, and Takuma Seno.
\newblock Simba: Simplicity bias for scaling up parameters in deep reinforcement learning.
\newblock {\em arXiv preprint arXiv:2410.09754}, 2024.

\bibitem{li2023near}
Donghao Li, Ruiquan Huang, Cong Shen, and Jing Yang.
\newblock Near-optimal conservative exploration in reinforcement learning under episode-wise constraints.
\newblock In {\em International Conference on Machine Learning}, pages 19527--19564. PMLR, 2023.

\bibitem{li2023reinforcement}
Shengbo~Eben Li.
\newblock Reinforcement learning for sequential decision and optimal control.
\newblock 2023.

\bibitem{lillicrap2015continuous}
Timothy~P Lillicrap, Jonathan~J Hunt, Alexander Pritzel, Nicolas Heess, Tom Erez, Yuval Tassa, David Silver, and Daan Wierstra.
\newblock Continuous control with deep reinforcement learning.
\newblock {\em arXiv preprint arXiv:1509.02971}, 2015.

\bibitem{lipman2022flow}
Yaron Lipman, Ricky~TQ Chen, Heli Ben-Hamu, Maximilian Nickel, and Matt Le.
\newblock Flow matching for generative modeling.
\newblock {\em arXiv preprint arXiv:2210.02747}, 2022.

\bibitem{lu2023synthetic}
Cong Lu, Philip Ball, Yee~Whye Teh, and Jack Parker-Holder.
\newblock Synthetic experience replay.
\newblock {\em Advances in Neural Information Processing Systems}, 36:46323--46344, 2023.

\bibitem{mnih2016asynchronous}
Volodymyr Mnih, Adria~Puigdomenech Badia, Mehdi Mirza, Alex Graves, Timothy Lillicrap, Tim Harley, David Silver, and Koray Kavukcuoglu.
\newblock Asynchronous methods for deep reinforcement learning.
\newblock In {\em International conference on machine learning}, pages 1928--1937. PmLR, 2016.

\bibitem{mnihHumanlevel2015}
Volodymyr Mnih, Koray Kavukcuoglu, David Silver, Andrei~A. Rusu, Joel Veness, Marc~G. Bellemare, Alex Graves, Martin Riedmiller, Andreas~K. Fidjeland, and Georg Ostrovski.
\newblock Human-level control through deep reinforcement learning.
\newblock 518(7540):529--533.

\bibitem{nauman2024bigger}
Michal Nauman, Mateusz Ostaszewski, Krzysztof Jankowski, Piotr Mi{\l}o{\'s}, and Marek Cygan.
\newblock Bigger, regularized, optimistic: scaling for compute and sample efficient continuous control.
\newblock {\em Advances in neural information processing systems}, 37:113038--113071, 2024.

\bibitem{neu2017unified}
Gergely Neu, Anders Jonsson, and Vicen{\c{c}} G{\'o}mez.
\newblock A unified view of entropy-regularized markov decision processes.
\newblock {\em arXiv preprint arXiv:1705.07798}, 2017.

\bibitem{osbandDeep2016}
Ian Osband, Charles Blundell, Alexander Pritzel, and Benjamin Van~Roy.
\newblock Deep exploration via bootstrapped dqn.
\newblock In {\em Proceedings of the International Conference on Machine Learning (ICML)}, 2016.

\bibitem{ostrovskiCountBased2017}
Georg Ostrovski, Marc~G. Bellemare, Aaron van~den Oord, and Remi Munos.
\newblock Count-based exploration with neural density models.
\newblock In {\em Advances in Neural Information Processing Systems (NeurIPS)}, 2017.

\bibitem{pathakCuriosityDriven2017}
Deepak Pathak, Pulkit Agrawal, Alexei~A. Efros, and Trevor Darrell.
\newblock Curiosity-driven exploration by self-supervised prediction.
\newblock In {\em Proceedings of the IEEE Conference on Computer Vision and Pattern Recognition (CVPR)}, 2017.

\bibitem{plappert2017parameter}
Matthias Plappert, Rein Houthooft, Prafulla Dhariwal, Szymon Sidor, Richard~Y Chen, Xi~Chen, Tamim Asfour, Pieter Abbeel, and Marcin Andrychowicz.
\newblock Parameter space noise for exploration.
\newblock {\em arXiv preprint arXiv:1706.01905}, 2017.

\bibitem{ramicicEntropybased2017}
Mirza Ramicic and Andrea Bonarini.
\newblock Entropy-based prioritized sampling in {{Deep Q-learning}}.
\newblock In {\em 2017 2nd {{International Conference}} on {{Image}}, {{Vision}} and {{Computing}} ({{ICIVC}})}, pages 1068--1072.

\bibitem{schaulPrioritized2016}
Tom Schaul, John Quan, Ioannis Antonoglou, and David Silver.
\newblock Prioritized experience replay.
\newblock In Yoshua Bengio and Yann LeCun, editors, {\em 4th International Conference on Learning Representations, {ICLR} 2016, San Juan, Puerto Rico, May 2-4, 2016, Conference Track Proceedings}, 2016.

\bibitem{schrittwieser2020mastering}
Julian Schrittwieser, Ioannis Antonoglou, Thomas Hubert, Karen Simonyan, Laurent Sifre, Simon Schmitt, Arthur Guez, Edward Lockhart, Demis Hassabis, Thore Graepel, et~al.
\newblock Mastering atari, go, chess and shogi by planning with a learned model.
\newblock {\em Nature}, 588(7839):604--609, 2020.

\bibitem{schulman2015trust}
John Schulman, Sergey Levine, Pieter Abbeel, Michael Jordan, and Philipp Moritz.
\newblock Trust region policy optimization.
\newblock In {\em International conference on machine learning}, pages 1889--1897. PMLR, 2015.

\bibitem{schulmanProximal2017}
John Schulman, Filip Wolski, Prafulla Dhariwal, Alec Radford, and Oleg Klimov.
\newblock Proximal {{Policy Optimization Algorithms}}, 2017.

\bibitem{sekar2020planning}
Ramanan Sekar, Oleh Rybkin, Kostas Daniilidis, Pieter Abbeel, Danijar Hafner, and Deepak Pathak.
\newblock Planning to explore via self-supervised world models.
\newblock In {\em International conference on machine learning}, pages 8583--8592. PMLR, 2020.

\bibitem{shin2017continual}
Hanul Shin, Jung~Kwon Lee, Jaehong Kim, and Jiwon Kim.
\newblock Continual learning with deep generative replay.
\newblock {\em Advances in neural information processing systems}, 30, 2017.

\bibitem{sohl2015deep}
Jascha Sohl-Dickstein, Eric Weiss, Niru Maheswaranathan, and Surya Ganguli.
\newblock Deep unsupervised learning using nonequilibrium thermodynamics.
\newblock In {\em International conference on machine learning}, pages 2256--2265. pmlr, 2015.

\bibitem{song2019generative}
Yang Song and Stefano Ermon.
\newblock Generative modeling by estimating gradients of the data distribution.
\newblock {\em Advances in neural information processing systems}, 32, 2019.

\bibitem{song2020score}
Yang Song, Jascha Sohl-Dickstein, Diederik~P Kingma, Abhishek Kumar, Stefano Ermon, and Ben Poole.
\newblock Score-based generative modeling through stochastic differential equations.
\newblock {\em arXiv preprint arXiv:2011.13456}, 2020.

\bibitem{sujitPrioritizing2023}
Shivakanth Sujit, Somjit Nath, Pedro H.~M. Braga, and Samira~Ebrahimi Kahou.
\newblock Prioritizing samples in reinforcement learning with reducible loss.
\newblock In Alice Oh, Tristan Naumann, Amir Globerson, Kate Saenko, Moritz Hardt, and Sergey Levine, editors, {\em Advances in Neural Information Processing Systems 36: Annual Conference on Neural Information Processing Systems 2023, NeurIPS 2023, New Orleans, LA, USA, December 10 - 16, 2023}, 2023.

\bibitem{sukhija2024maxinforl}
Bhavya Sukhija, Stelian Coros, Andreas Krause, Pieter Abbeel, and Carmelo Sferrazza.
\newblock Maxinforl: Boosting exploration in reinforcement learning through information gain maximization.
\newblock {\em arXiv preprint arXiv:2412.12098}, 2024.

\bibitem{sukhija2025optimism}
Bhavya Sukhija, Lenart Treven, Carmelo Sferrazza, Florian Dorfler, Pieter Abbeel, and Andreas Krause.
\newblock Optimism via intrinsic rewards: Scalable and principled exploration for model-based reinforcement learning.
\newblock In {\em 7th Robot Learning Workshop: Towards Robots with Human-Level Abilities}, 2025.

\bibitem{sutton1998reinforcement}
Richard~S Sutton, Andrew~G Barto, et~al.
\newblock {\em Reinforcement learning: An introduction}, volume~1.
\newblock MIT press Cambridge, 1998.

\bibitem{tassa2018deepmind}
Yuval Tassa, Yotam Doron, Alistair Muldal, Tom Erez, Yazhe Li, Diego de~Las Casas, David Budden, Abbas Abdolmaleki, Josh Merel, Andrew Lefrancq, et~al.
\newblock Deepmind control suite.
\newblock {\em arXiv preprint arXiv:1801.00690}, 2018.

\bibitem{varakantham2023constrained}
Pankayaraj Pathmanathan~Pradeep Varakantham.
\newblock Constrained reinforcement learning in hard exploration problems.
\newblock In {\em AAAI Conference on Artificial Intelligence (AAAI)}, 2023.

\bibitem{vaswani2017attention}
Ashish Vaswani, Noam Shazeer, Niki Parmar, Jakob Uszkoreit, Llion Jones, Aidan~N Gomez, {\L}ukasz Kaiser, and Illia Polosukhin.
\newblock Attention is all you need.
\newblock {\em Advances in neural information processing systems}, 30, 2017.

\bibitem{wang2025deep}
Kevin Wang, Hongqian Niu, Yixin Wang, and Didong Li.
\newblock Deep generative models: Complexity, dimensionality, and approximation.
\newblock {\em arXiv preprint arXiv:2504.00820}, 2025.

\bibitem{wang2024prioritized}
Renhao Wang, Kevin Frans, Pieter Abbeel, Sergey Levine, and Alexei~A Efros.
\newblock Prioritized generative replay.
\newblock {\em arXiv preprint arXiv:2410.18082}, 2024.

\bibitem{wang2024reinforcement}
Shuhe Wang, Shengyu Zhang, Jie Zhang, Runyi Hu, Xiaoya Li, Tianwei Zhang, Jiwei Li, Fei Wu, Guoyin Wang, and Eduard Hovy.
\newblock Reinforcement learning enhanced llms: A survey.
\newblock {\em arXiv preprint arXiv:2412.10400}, 2024.

\bibitem{wang2024diffusion}
Yinuo Wang, Likun Wang, Yuxuan Jiang, Wenjun Zou, Tong Liu, Xujie Song, Wenxuan Wang, Liming Xiao, Jiang Wu, Jingliang Duan, et~al.
\newblock Diffusion actor-critic with entropy regulator.
\newblock {\em Advances in Neural Information Processing Systems}, 37:54183--54204, 2024.

\bibitem{wang2025enhanced}
Yinuo Wang, Likun Wang, Mining Tan, Wenjun Zou, Xujie Song, Wenxuan Wang, Tong Liu, Guojian Zhan, Tianze Zhu, Shiqi Liu, et~al.
\newblock Enhanced dacer algorithm with high diffusion efficiency.
\newblock {\em arXiv preprint arXiv:2505.23426}, 2025.

\bibitem{zhangSTORM2023}
Weipu Zhang, Gang Wang, Jian Sun, Yetian Yuan, and Gao Huang.
\newblock {STORM:} efficient stochastic transformer based world models for reinforcement learning.
\newblock In Alice Oh, Tristan Naumann, Amir Globerson, Kate Saenko, Moritz Hardt, and Sergey Levine, editors, {\em Advances in Neural Information Processing Systems 36: Annual Conference on Neural Information Processing Systems 2023, NeurIPS 2023, New Orleans, LA, USA, December 10 - 16, 2023}, 2023.

\end{thebibliography}

\newpage
\appendix

\section{Theoretical Analysis}
\label{app:thm}
\subsection{Useful lemmas }

\begin{lemma}[Discounted‑Occupancy Lipschitz Lemma]\label{lem:lipschitz}
Let  $0<\gamma<1$.
For any two stochastic policies $\pi$ and $\pi'$ sharing the same initial state distribution $\mu$, their discounted state–occupancy measurements
$d_\mu^{\pi}(s)$ and $d_\mu^{\pi'}(s)$ meet the following condition:
\begin{equation}
\bigl\|d_\mu^{\pi}-d_\mu^{\pi'}\bigr\|_{\mathrm{TV}}
\;\le\;
\frac{\gamma}{1-\gamma}\;
\sup_{s\in\mathcal{S}}
\operatorname{TV}\bigl(\pi(\cdot\mid s),\pi'(\cdot\mid s)\bigr),
\end{equation}
where $TV(\cdot,\cdot)$ denotes the total variation of two distributions:
\begin{equation}
    TV(a(s),b(s))=\frac{1}{2}\sum_s|a(s)-b(s)|.
\end{equation}
\end{lemma}

\begin{proof}
By the definition, the discounted occupancy measure can be reformulated as a fixed-point equation:
\begin{equation}
d_\mu^{\pi}
=(1-\gamma)\mu+\gamma\,d_\mu^{\pi}P_\pi,
\quad
d_\mu^{\pi'}
=(1-\gamma)\mu+\gamma\,d_\mu^{\pi'}P_{\pi'} ,
\end{equation}

where $P_\pi=\sum_{a}\pi(a\mid s)P(s'\mid s, a)$ is the one-step transition kernel of $\pi$. It denotes the transition probability from state $s$ to $s'$ under the policy $\pi$. Defining the subtraction of the two transition kernels as $\Delta$, the following equation holds:
\begin{equation}
\Delta
:=d_\mu^{\pi}-d_\mu^{\pi'}
=\gamma\,\Delta P_\pi
+\gamma\,d_\mu^{\pi'}(P_\pi-P_{\pi'}).
\end{equation}
By rearranging $\Delta$ to the left-hand side and factoring out the common terms, we obtain:
\begin{equation}\Delta(I-\gamma P_\pi)=\gamma d_\mu^{\pi^{\prime}}(P_\pi-P_{\pi^{\prime}}).\end{equation}
Because the induced norm of the transition kernel $\|P_\pi\|_{1}=1$ and $\gamma<1$, $(I-\gamma P_\pi)^{-1}$ exists and
$(I-\gamma P_\pi)^{-1}=\sum_{k\ge0}(\gamma P_\pi)^k$ converges.
Multiplying by this inverse, we can derive the Neumann series as follows:
\begin{equation}
\Delta
=\gamma\sum_{k=0}^{\infty}(\gamma P_\pi)^k\,
       d_\mu^{\pi'}(P_\pi-P_{\pi'}).
\label{eq:Neumann}
\end{equation}
The $L_1$ norm of the above equation is given by:
\begin{equation}
\|\Delta\|_{1}
\le
\gamma
\Bigl(\sum_{k=0}^{\infty}\gamma^{k}\Bigr)
\|d_\mu^{\pi'}\|_{1}
\|P_\pi-P_{\pi'}\|_{1}.
\end{equation}
Due to the definition of the transition kernel, we can derive that:
\begin{equation}
\begin{aligned}
    \left\|P_\pi-P_{\pi^{\prime}}\right\|_{1\to1}&=\sup_{s\in\mathcal{S}}\left\|\sum_{a\in\mathcal{A}}[\pi(a\mid s)-\pi^{\prime}(a\mid s)]P(\cdot\mid s,a)\right\|_1\\
    &\leq\sum_{a\in\mathcal{A}}\lvert\pi(a\mid s)-\pi^{\prime}(a\mid s)\rvert\cdot||1||\\
    &=2\operatorname{TV}(\pi(\cdot\mid s),\pi'(\cdot\mid s)\bigr)
\end{aligned}
\end{equation}
Since $\|d_\mu^{\pi'}\|_{1}=1$, We can obtain that:
\begin{equation}
\|\Delta\|_{1}\le
\frac{2\gamma}{1-\gamma}\sup_{s}\operatorname{TV}(\pi(\cdot\mid s),\pi'(\cdot\mid s)\bigr).
\end{equation}
Dividing by 2 converts the $L_1$ norm to total variation, the proof is completed.

For clarity, we provide the proof in the discrete state-action space. Since we assume the continuous state-action space is a measurable continuous space, the contraction property in the total variation norm and the convergence of the Neumann series of $(I - \gamma P)$ also hold in this setting. Therefore, the conclusion can be directly extended to the continuous state-action space.
\end{proof}

\begin{lemma}[FIFO‑Buffer Proximity Lemma]\label{lem:buffer}
Assume the replay buffer $\mathcal{B}$ holds exactly the most recent $K$ transitions
(\emph{FIFO} of fixed size~$K$).
We can define the occupancy measurement under the behavior policy in $\mathcal{B}$ as:
$
\nu_t
=\frac1K\sum_{h=0}^{K-1}d^{\pi_{t-h}}$,
and let $
\Delta_\tau
:=\sup_{s\in\mathcal{S}}
    \operatorname{TV}\bigl(\pi_{\tau+1}(\cdot\mid s),\pi_\tau(\cdot\mid s)\bigr).
$
Then the following inequality holds:
\begin{equation}
\bigl\|\nu_t-d^{\pi_t}\bigr\|_{\mathrm{TV}}
\;\le\;
\frac{\gamma}{1-\gamma}\,
\sum_{j=0}^{K-1}\Delta_{t-j}.
\end{equation}
\end{lemma}

\begin{proof}
By the convexity of total variation, we can obtain:
\begin{equation}
\|\nu_t-d^{\pi_t}\|_{\mathrm{TV}}
\le
\frac1K\sum_{h=0}^{K-1}
     \|d^{\pi_{t-h}}-d^{\pi_t}\|_{\mathrm{TV}}.
\end{equation}
By using the triangle inequality in, we can obtain that
$
\|d^{\pi_{t-h}}-d^{\pi_t}\|_{\mathrm{TV}}
 \le
 \tau_{\text{mix}}\sum_{j=0}^{h-1}\Delta_{t-j}
$
with $\tau_{\text{mix}}=\gamma/(1-\gamma)$.
Insert this bound and interchange the sums, we can derive that:
\begin{equation}
\begin{aligned}
\|\nu_t-d^{\pi_t}\|_{\mathrm{TV}}
&\le
\frac{\tau_{\text{mix}}}{K}
\sum_{h=0}^{K-1}\sum_{j=0}^{h-1}\Delta_{t-j}
\\
&=
\tau_{\text{mix}}
\sum_{j=0}^{K-1}
\frac{K-j}{K}\,\Delta_{t-j}
\;\le\;
\tau_{\text{mix}}\sum_{j=0}^{K-1}\Delta_{t-j}.
\end{aligned}
\end{equation}
\end{proof}

\begin{lemma}[Behaviour‑Mixing Contraction Lemma]\label{lem:beta}
Assuming that each policy $\pi_{t-h}$ (for $h \in [0, K-1]$) writes the same number of transition samples into the buffer during the last $K$ time steps, and the difference between consecutive policies remains at a negligible scale: $\sup_s \operatorname{TV}(\pi_{t+1}, \pi_t)(s) \ll 1$, the behavior policy represented in the buffer can be approximately expressed as $\beta_t\approx \frac{1}{K} \sum_{h=0}^{K-1} \pi_{t-h}$
Let $\delta_\tau
:=\sup_{s\in\mathcal{S}}
     \operatorname{TV}\bigl(\pi_\tau(\cdot\mid s),\pi^{*}(\cdot\mid s)\bigr)$, then the following bound holds:
\begin{equation}
\sup_{s\in\mathcal{S}}
\operatorname{TV}\bigl(\beta_t(\cdot\mid s),\pi^{*}(\cdot\mid s)\bigr)
\;\le\;
\frac1K\sum_{h=0}^{K-1}\delta_{\,t-h}.
\end{equation}

\end{lemma}

\begin{proof}
Fix $s\in\mathcal{S}$, since the total variation is convex in its first argument, we can derive that:
\begin{equation}
\operatorname{TV}\Bigl(
        \frac1K\sum_{h=0}^{K-1}\pi_{t-h}(\cdot\mid s),
        \pi^{*}(\cdot\mid s)\Bigr)
\le
\frac1K\sum_{h=0}^{K-1}
     \operatorname{TV}\bigl(\pi_{t-h}(\cdot\mid s),\pi^{*}(\cdot\mid s)\bigr).
\end{equation}
Taking the supremum over $s$ and moving the $\sup$ outside the sum gives the claimed inequality.
\end{proof}

\begin{lemma}[The $\lambda$–mixture estimator bias]
\label{lem:lambda}
Let $\rho(s,a)=d_{\textbf{env}}^{\pi}(s,a)$ be the target state–action
distribution and $\rho'(s,a)=d_{\text{gen}}^{\beta}(s,a)$ be the behaviour distribution.
For any measurable function $g:\mathcal S\times\mathcal A\to\mathbb R$
bounded by $\lVert g\rVert_\infty$,
define the $\lambda$–mixture sampling measure as
$q_\lambda =(1-\lambda)\rho +\lambda \rho',
0\le\lambda\le1$.
Denote the exact importance sampling (IS) estimator
$\hat J_{\text{IS}}=\mathbb E_{(s,a)\sim q_\lambda}\bigl[\,w(s,a)\,g(s,a)\bigr]$
with weight $w=\rho/q_\lambda$,
and the \emph{$\lambda$–mixture estimator}
$\hat J_{\text{mix}}
     = \mathbb E_{(s,a)\sim q_\lambda}\bigl[g(s,a)\bigr]$.
Then the following inequality holds:
\begin{equation}
    \bigl|\,
        \mathbb E\bigl[\hat J_{\text{mix}}\bigr]
        -\mathbb E\bigl[\hat J_{\text{IS}}\bigr]\bigr|\le
    \lambda\,M\,\sqrt{2\,
           \mathrm{KL}\bigl(\rho'\,\Vert\,\rho\bigr)}=
    \mathcal O\bigl(
        \lambda\,\mathrm{KL}(\pi_\theta\Vert\beta)^{1/2}
    \bigr).
\end{equation}

\begin{proof}
Rewrite the bias as follows:
\begin{equation}
\hat J_{\text{IS}}-\hat J_{\text{mix}}
 =\lambda\,\mathbb E_{\rho'}[\,\bigl(\tfrac{\rho}{\rho'}-1\bigr)\,g\,].
\end{equation}
By applying Hölder’s inequality, we can obtain that:
\begin{equation}
\lVert \rho-\rho'\rVert_1
     \le\sqrt{2\,\mathrm{KL}(\rho'\Vert \rho)}.
\end{equation}
Because $|g|\le M$, the magnitude of the bias is upper‑bounded by $\lambda\,M\sqrt{2\,\mathrm{KL}(\rho'\Vert \rho)}$.
Replacing the distribution notation with the corresponding policies gives the stated $\mathcal O(\lambda\,\mathrm{KL})$ dependence.
\end{proof}
\end{lemma}

\subsection{Proof of Theorem 1}
\label{proof 1}
\begin{proof}
From lemma \ref{lem:beta}, the upper bound of total variation between the behavior policy $\beta_t$ and the optimal policy $\pi^{*}$ can be derived as:
\begin{equation}
    \sup_s\mathrm{TV}(\beta_t,\pi^{*})\leq\frac{1}{K}\sum_{h=0}^{K-1}\delta_{t-h}.
\end{equation}
If the target policy converges to the optimal policy, i.e., $\delta_{t-h}\to 0$  holds, then the following relation is satisfied:
\begin{equation}\lim_{t\to\infty}\sup_s\mathrm{TV}(\beta_t,\pi^{*})=0.\end{equation}
The above expression indicates that the behavior policy $\beta_t$ will gradually converge to the optimal policy $\pi^*$. Then, using the triangle inequality, we can obtain the following:
\begin{equation}\mathrm{TV}(\nu_t,d^{\pi^*})\leq\underbrace{\mathrm{TV}(\nu_t,d^{\pi_t})}_{A_t}+\underbrace{\mathrm{TV}(d^{\pi_t},d^{\pi^*})}_{B_t}.\end{equation}

From lemma \ref{lem:buffer}, the first part $A_t$ can be derived as :
\begin{equation}A_t\leq\tau_{\min}\sum_{j=0}^{K-1}\Delta_{t-j}.\end{equation}
Since the policy can converge to the optimal policy within a finite number of steps $N$, the final policy shift error is considered to vanish after these steps. Therefore, the term $A_t$ can be regarded as zero. From lemma \ref{lem:lipschitz}, the $B_t$ can be reformulated as:
\begin{equation}B_t=\left\|d^{\pi_t}-d^{\pi^*}\right\|_{\mathrm{TV}}\leq\tau_{\mathrm{mix}}\sup_s\mathrm{TV}\left(\pi_t(\cdot|s),\pi^{\cdot^*}(\cdot|s)\right)=\tau_{\mathrm{mix}}\delta_t.\end{equation}
Likewise, due to policy convergence in finite steps, $\delta_t=0$ holds when $t>N$. Thus, the proof is complete.
\end{proof}
Notably, if the policy does not converge to the optimal policy within a finite number of steps, but the policy shift $\Delta_t$ decays sufficiently fast after each update, where the following condition holds:
\begin{equation}
\sum_{t=0}^\infty\Delta_t=\sum_{t=0}^\infty\sup_{s\in S}\mathrm{TV}(\pi_{t+1}(\cdot\mid s),\pi_t(\cdot\mid s))<\infty.
\end{equation}
Under this condition, it can still be guaranteed that $A_t=0$, and the original proof remains valid.

\subsection{Discussion of importance sampling in MoGE}
\label{discussion 1}
In MoGE, since the critical experiences are generated by the diffusion-based generator and the one-step imagination world model, both the initial‑state distribution and the policy of the experiences may differ from those in replay buffer $\mathcal{B}$ under the assumption that the one-step world model can accurately estimate the environmental dynamics, where all transitions share the same dynamics kernel $P(s' \mid s, a)$.

\textbf{Problem Setting.} We consider training samples composed of two distributions: \textbf{(1)}$\mathcal D_{\text{env}}$ collected with the behavior policy
$\beta$ and the real initial state distribution $d_{\text{env}}(s)$
from the buffer $\mathcal{B}$;
\textbf{(2)} $\mathcal D_{\text{gen}}$ generated synthetically with target policy $\pi_\theta$ by MoGE, where the initial state distribution is denoted as $d_{\text{gen}}(s)$.

For policy evaluation, let $Q^\pi$ be the action-value function of the target policy
$\pi_\theta$.  It obeys the Bellman identity
\begin{equation}
  Q^\pi(s,a) \;=\;
  \mathbb E_{s' \sim P,\,a' \sim \pi_\theta}
  \bigl[r(s,a) + \gamma\,Q^\pi(s',a')\bigr],
  \qquad \forall (s,a).
  \label{eq:bellman}
\end{equation}
Because Eq.~\eqref{eq:bellman} holds in a point-wise way, where the $(s,a)$ can be sampled from anonymous distribution, the squared TD--error of any parameterised critic $Q_\psi$,
\begin{equation}
  \mathcal L_{\text{PEV}}(\psi)=
  \mathbb E_{(s,a,r,s') \sim \mathcal D_k}
  \Bigl[
      Q_\psi(s,a) -
      \bigl(r + \gamma\,\mathbb E_{a' \sim \pi_\theta}
            [Q_{\bar \psi}(s',a')]\bigr)
  \Bigr]^2,
  \label{eq:critic-loss}
\end{equation}
still attains its global minimum at $Q_\phi = Q^\pi$ for
any mixture of the training $\mathcal D_{k}=(1-k)\mathcal D_{\text{env}} + k\mathcal D_{\text{gen}}, 0 \le k \le 1.$
Hence, the critic remains unbiased without importance sampling; off-policy sampling only affects optimization
variance.  An IS correction becomes necessary only when regressing the full Monte-Carlo return $G_0$, which is
behaviour-dependent. However, for policy improvement, the importance sampling is non-negligible since the existence of a distribution mismatch:
\begin{equation}
  \mathcal L_{\text{PIM}}(\theta)=\mathbb{E}_{(s,a)\,\sim\,\mathcal D_{\text{env}}}
  \Bigl[
      \rho(s,a)\;
      g(s,a)
  \Bigr]
+
  \mathbb{E}_{(s,a)\,\sim\,\mathcal D_{\text{gen}}}
  \Bigl[
      w(s)\;
      g(s,a)
  \Bigr]
\end{equation}
where $g(s,a)$ denote the return differs in different algorithm, and $w(s)=\frac{d_{\text{env}}(s)}{d_{\text{gen}}(s)}, \rho(s,a)=\frac{\pi_\theta(a\mid s)}{\beta(a\mid s)}$. However, for algorithms that compute the objective function directly based on the target, i.e., $g(s,a) = g(s, \pi_\theta(s))$, importance sampling for the policy can be omitted. In this case, policy improvement only requires importance sampling for the initial state distribution.
Since computing $w(s)$ and is intractable in high-dimensional
continuous spaces.  We therefore approximate the IS expectation by a sampling mixture using a mixing rate $\lambda$ since the bias can be bound by Lemma \ref{lem:lambda}. The target for policy improvement in MoGE can be finally derived as a \emph{sampling mixture}:
\begin{equation}
  \mathcal L_{\text{PIM}}(\theta)
  =(1-\lambda)\mathbb{E}_{(s,a)\,\sim\,\mathcal D_{\text{env}}}
  \Bigl[
      g(s,a)
  \Bigr]
+
  \lambda\mathbb{E}_{(s,a)\,\sim\,\mathcal D_{\text{gen}}}
  \Bigl[
      g(s,a)
  \Bigr]
\end{equation}
with $\lambda \in [0,1)$. For example, the resulting actor loss for SAC-style objectives is like:
\begin{align}
  \mathcal L_{\text{PIM}}(\theta)
  &=
  (1-\lambda)\,
  \mathbb E_{(s,a)\sim\mathcal D_{\text{env}}}
   \bigl[\alpha\log\pi_\theta(a\mid s)-Q_\psi(s,a)\bigr]
  \nonumber\\
  &\quad+
  \lambda\,
  \mathbb E_{(s,a)\sim\mathcal D_{\text{beh}}}
   \bigl[\alpha\log\pi_\theta(a\mid s)-Q_\psi(s,a)\bigr].
  \label{eq:actor-loss}
\end{align}

Note that Eq. \ref{eq:actor-loss} is the first-order Taylor expansion of the exact IS estimator; its bias scales with
$\mathcal O\bigl(\lambda\,\mathrm{KL}(\pi_\theta\Vert\beta)^{1/2}\bigr)$.
Empirically, a small $\lambda$ retains the coverage benefit of the generated data while keeping bias and training instability negligible.

Theoretically, the choice of $k$ is unrestricted; however, to reduce training variance, it is advisable to constrain $k$ within a smaller range. In this work, we set $k = 2\lambda$.

\section{Environmental configuration}
\label{app: details}
\subsection{Environment Introduction}
\label{environment}
\textbf{DeepMind Control Suite}. We chose 5 challenging tasks involving the humanoid and quadruped robots. The final reward for each task is the product of the standing reward and the forward velocity reward, expressed as:
$\textbf{\textit{Reward}} = (\textbf{\textit{Standing Reward}}) \times (\textbf{\textit{Forward Velocity Reward}})$.
\begin{figure}[h!]
        \centering
        \begin{subfigure}[b]{0.32\textwidth}
        \includegraphics[width=\linewidth]{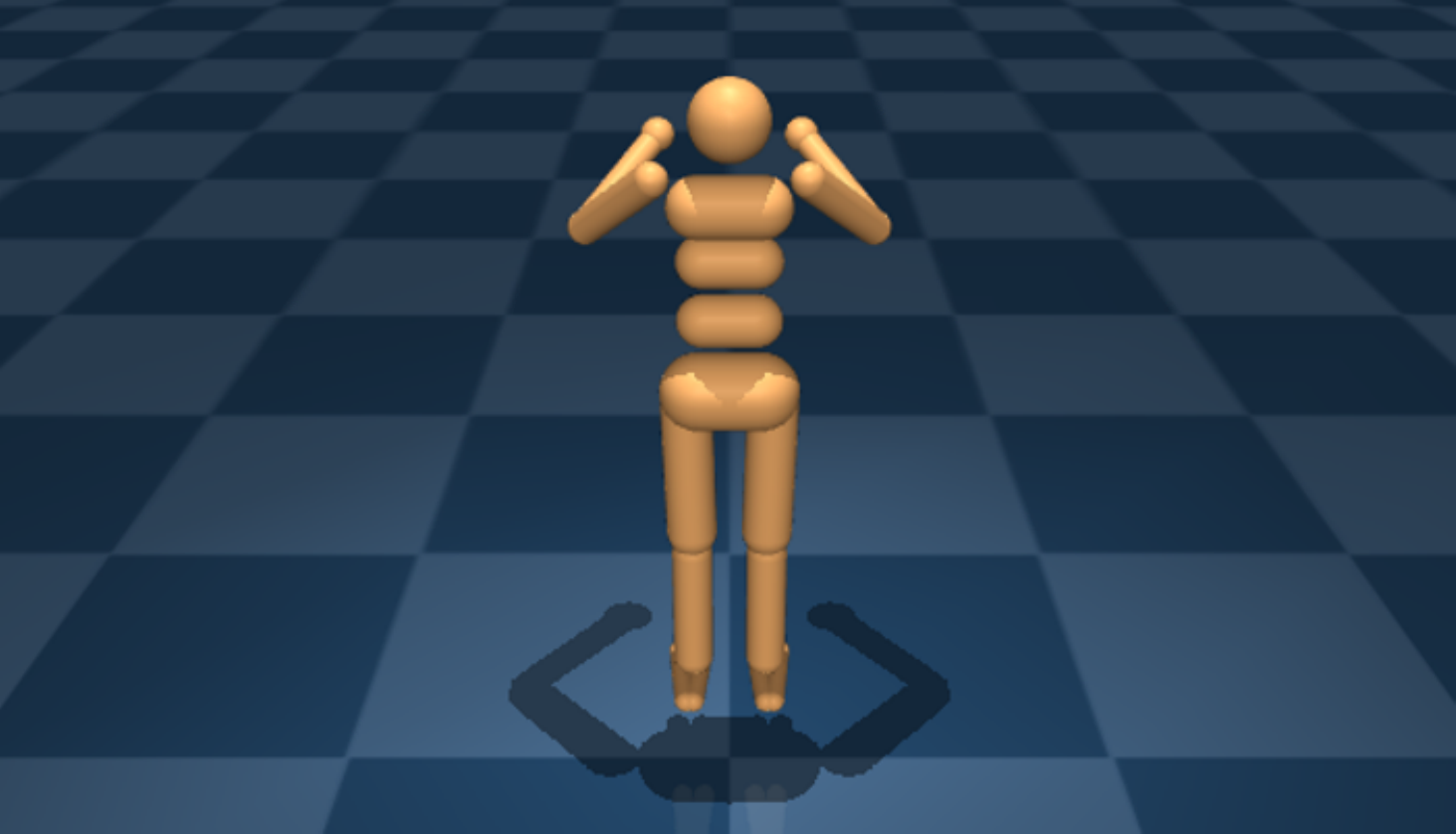}
        \caption{Humanoid}
        \label{fig:humanoid}
        \end{subfigure}
        \begin{subfigure}[b]{0.32\textwidth}
        \includegraphics[width=\linewidth]{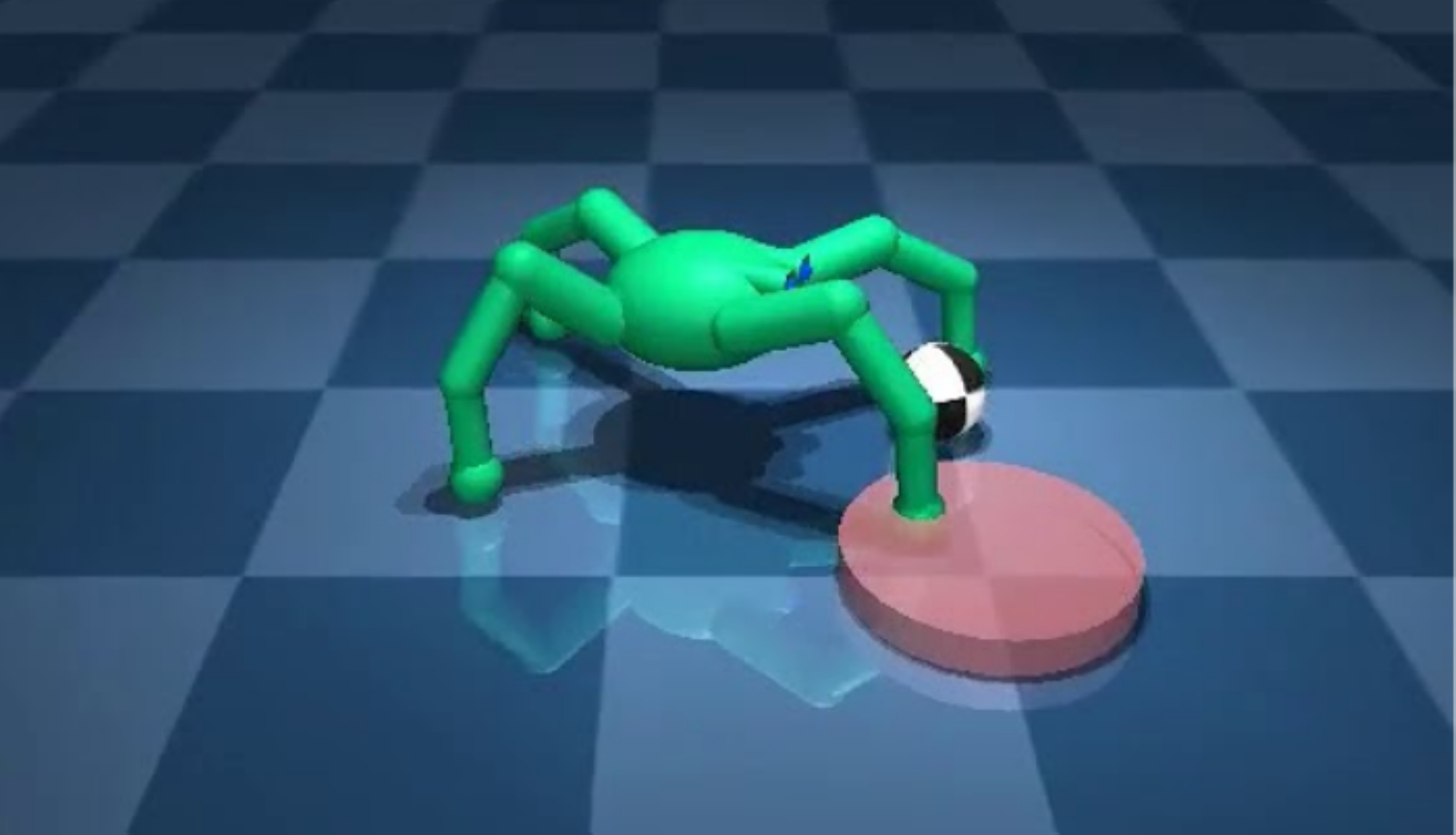}
        \caption{Quadruped}
        \label{fig:quadruped}
        \end{subfigure}
    \caption{DMC environments.}
\end{figure}

\textit{Humanoid tasks}: The Humanoid consists of three primary tasks, each designed to challenge an agent's ability to control a simulated humanoid robot. The three tasks are described as follows:

\begin{itemize}
    \item \textbf{Stand}: The agent's objective is to maintain an upright posture. The reward function encourages stability and a vertical torso position while minimizing deviations from an ideal standing height.
    \item \textbf{Walk}: The agent is rewarded for moving forward at a target velocity of 1 m/s. This task evaluates the agent's capability to coordinate limb movement and maintain balance while walking.
    \item \textbf{Run}: In this task, the agent is required to achieve a high-speed forward motion of 10 m/s. The challenge includes maintaining dynamic stability and efficient stride patterns.
\end{itemize}

\textit{Quadruped tasks}: The Quadruped environment represents a four-legged robotic model with two major tasks aimed at testing multi-legged coordination and locomotion:

\begin{itemize}
    \item \textbf{Walk}: The agent must achieve forward movement at a steady pace. This task assesses the stability and synchronization of its four legs during controlled walking.
    \item \textbf{Run}: The agent is required to accelerate to higher velocities, demanding agile gait adjustments and robust stability during high-speed movement.
\end{itemize}

\textbf{OpenAI Gym}. We chose 5 widely used locomotion tasks in various domains:

\begin{figure}[h!]

    \begin{minipage}[b]{0.25\textwidth}
        \centering
        \includegraphics[width=\linewidth]{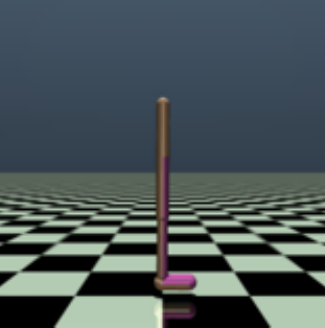}
        \caption{Walker2d-v3}
        \label{fig:walker2d}
    \end{minipage}
    \quad
    \begin{minipage}[b]{0.6\textwidth}
    \small
        \textbf{\textit{State-action space}}: $\mathcal{S} \in \mathbb{R}^{17}$, $\mathcal{A} \in \mathbb{R}^{6}$.\\

        \textbf{\textit{Objective.}} Maintain forward velocity as fast as possible while avoiding falling over.\\

        \textbf{\textit{Initialization.}} The walker is initialized in a standing position with slight random noise added to joint positions and velocities.\\

        \textbf{\textit{Termination.}} The episode ends when the agent falls, the head touches the ground, or after 1000 steps.\\
        \vspace{8pt}
    \end{minipage}
\end{figure}

\begin{figure}[h!]
\small
    \begin{minipage}[b]{0.25\textwidth}
        \centering
        \includegraphics[width=\linewidth]{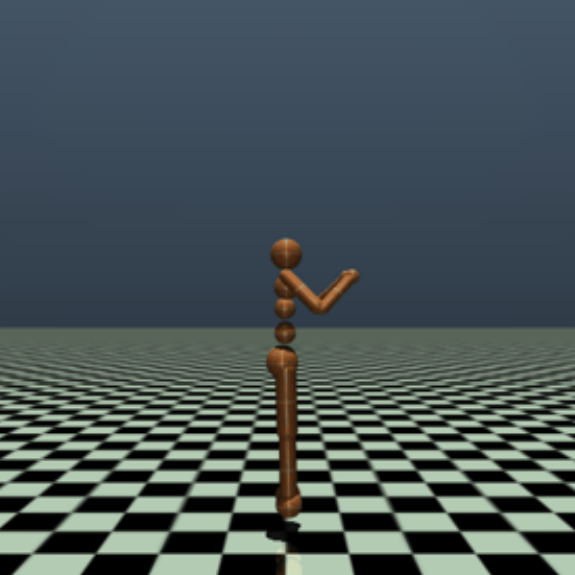}
        \caption{Humanoid-v3}
        \label{fig:humanoid}
    \end{minipage}
    \quad
    \begin{minipage}[b]{0.6\textwidth}
\textbf{\textit{State-action space}}: $\mathcal{S} \in \mathbb{R}^{376}$, $\mathcal{A} \in \mathbb{R}^{17}$.\\

        \textbf{\textit{Objective.}} Maintain balance and walk or run forward at a high velocity while avoiding falls.\\

\textbf{\textit{Initialization.}} The humanoid starts in an upright position with slight random perturbations to joint angles and velocities.\\

\textbf{\textit{Termination.}} The episode ends when the head height is less than 1.0 meter, the torso tilts excessively, or after 1000 steps.\\
        \vspace{8pt}
    \end{minipage}
\end{figure}

\begin{figure}[h!]
    \begin{minipage}[b]{0.25\textwidth}
        \centering
        \includegraphics[width=\linewidth]{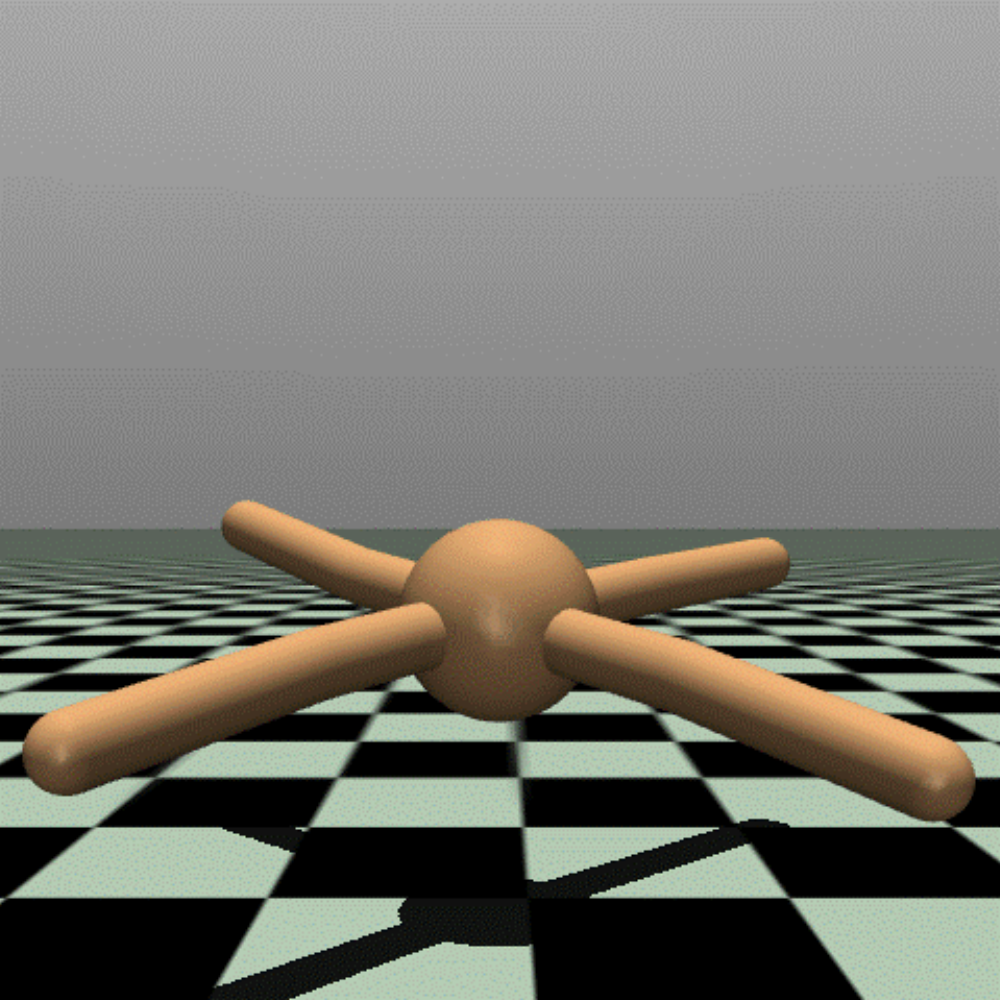}
        \caption{Ant-v3}
        \label{fig:ant}
    \end{minipage}
    \quad
    \begin{minipage}[b]{0.6\textwidth}
    \small
\textbf{\textit{State-action space}}: $\mathcal{S} \in \mathbb{R}^{111}$, $\mathcal{A} \in \mathbb{R}^{8}$.\\

\textbf{\textit{Objective.}} Navigate forward as quickly as possible using four legs while maintaining stability.\\

\textbf{\textit{Initialization.}} The ant is initialized in a stable, upright position with random noise applied to its joints.\\

\textbf{\textit{Termination.}} The episode ends if the ant falls, flips over, or reaches the maximum step count of 1000.\\
        \vspace{8pt}
    \end{minipage}
\end{figure}

\begin{figure}[h!]
    \begin{minipage}[b]{0.25\textwidth}
        \centering
        \includegraphics[width=\linewidth]{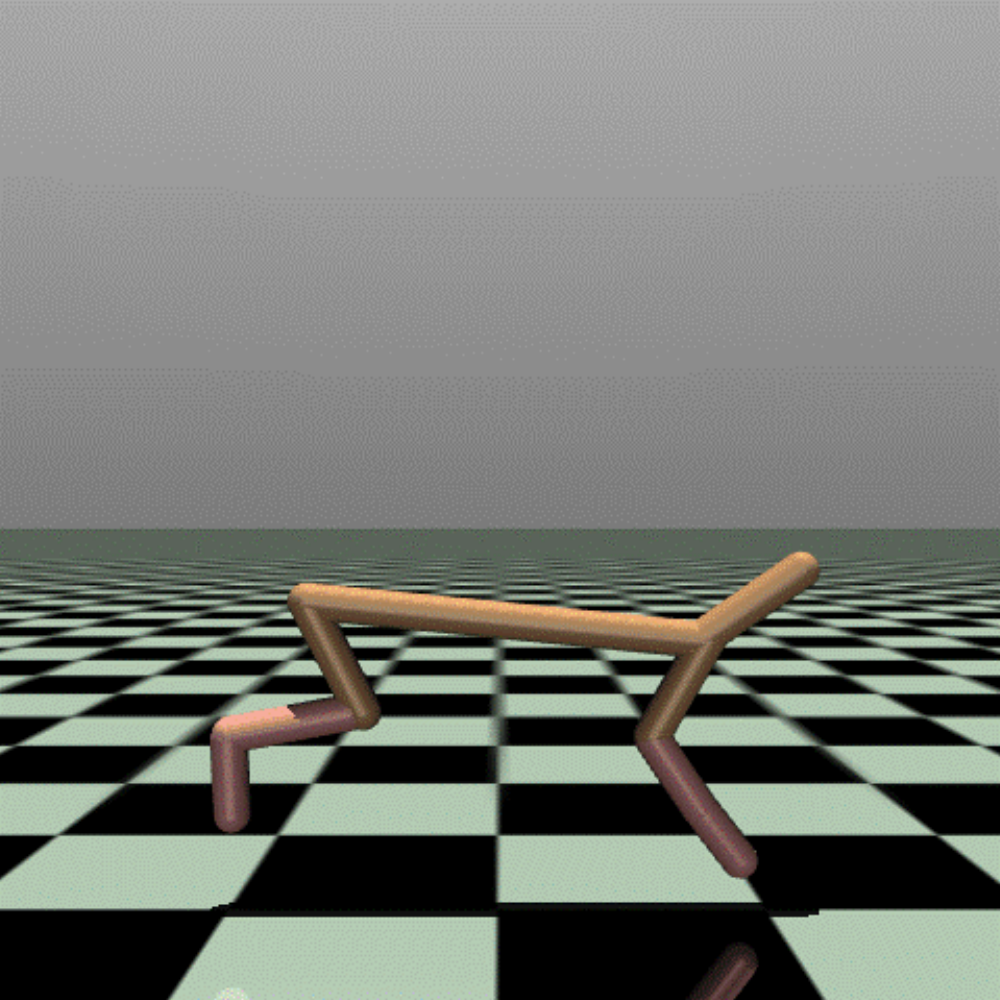}
        \caption{Halfcheetah-v3}
        \label{fig:halfcheetah}
    \end{minipage}
    \quad
    \begin{minipage}[b]{0.6\textwidth}
    \small
\textbf{\textit{State-action space}}: $\mathcal{S} \in \mathbb{R}^{17}$, $\mathcal{A} \in \mathbb{R}^{6}$.\\

\textbf{\textit{Objective.}} Achieve maximum forward velocity with smooth, coordinated movements.\\

\textbf{\textit{Initialization.}} The agent starts with a slight forward tilt and randomized joint noise.\\

\textbf{\textit{Termination.}} The episode ends after 1000 steps or if the agent's head touches the ground.\\
        \vspace{8pt}
    \end{minipage}
\end{figure}

\begin{figure}[h!]
    \begin{minipage}[b]{0.25\textwidth}
        \centering
        \includegraphics[width=\linewidth]{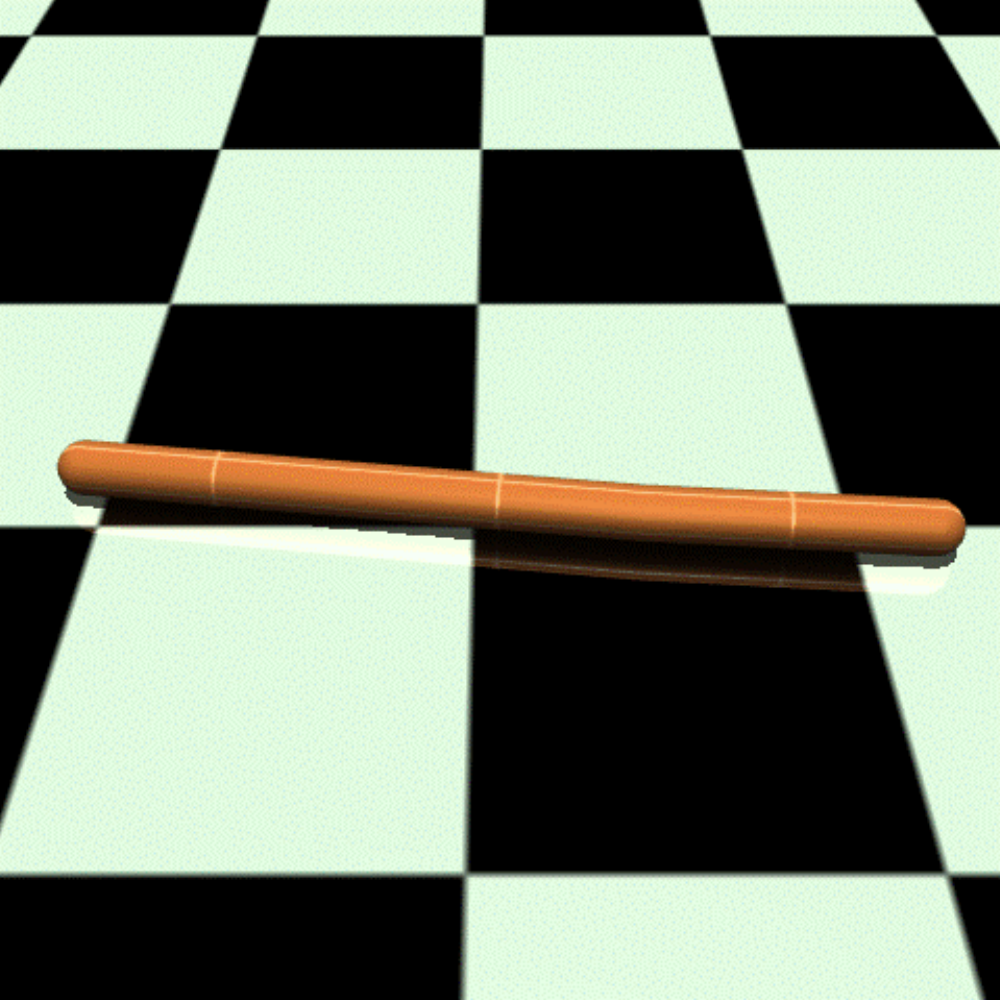}
        \caption{Swimmer-v3}
        \label{fig:swimmer}
    \end{minipage}
    \quad
    \begin{minipage}[b]{0.6\textwidth}
    \small
\textbf{\textit{State-action space}}: $\mathcal{S} \in \mathbb{R}^{8}$, $\mathcal{A} \in \mathbb{R}^{2}$.\\

\textbf{\textit{Objective.}} Propel forward through water-like dynamics using sinusoidal wave patterns.\\

\textbf{\textit{Initialization.}} The swimmer starts in a straight posture with minor random perturbations.\\

\textbf{\textit{Termination.}} The episode ends after 1000 steps, with no explicit termination for falling.\\
        \vspace{8pt}
    \end{minipage}
\end{figure}

\subsection{Reproducibility Statement \&  Detailed Hyperparameters}
\label{reproduce}
\label{hyperparameters}

In MoGE, we adopt the hyperparameter settings without additional fine-tuning and use the same configuration across all previously demonstrated tasks, which are listed in Table \ref{tab_hyperparams}. Our core algorithm file is accessible at \url{https://github.com/WangLK-Franklin/MoGE}.

In this paper, all experiments are conducted with a total of 1.5 million environment interaction steps, and the results are averaged over three random seeds. The experiments are performed on an AMD Ryzen Threadripper 3960X 24-Core Processor and an NVIDIA GeForce RTX 4090 GPU. Besides, the walltime results(s) of $1.5M$ steps are included in Table \ref{tab:moge_time}.

\begin{table}[t]
\centering
\fontsize{7pt}{9}\selectfont
\setlength{\tabcolsep}{4pt}
\caption{Performance comparison of MoGE and related baselines on Humanoid-run task.}
\begin{tabular}{lcccc}
\toprule
Env & MoGE & PGR & SER & DSAC \\
\midrule
Humanoid-run & 151657 $\pm$ 395 & 207813±1464 & 210321±788 & 128153±122 \\
\bottomrule
\end{tabular}
\label{tab:moge_time}
\end{table}
The results show that MoGE incurs slightly higher walltime compared to baseline algorithms. Notably, to reduce computational overhead, we perform MoGE training only once every 10 environment interaction steps, which significantly controls the overall compute cost.
Moreover, unlike methods requiring retraining the generative model on the buffer until convergence before use, MoGE continuously updates its generator during training. This makes MoGE’s diffusion-based training more efficient and lightweight in practice.
\begin{table}[h]
\caption{Hyperparameter settings. }
\label{tab_hyperparams}
\vskip 0.15in
\begin{center}
\begin{small}
\begin{tabular}{l l | l l}
\toprule
\textbf{Hyperparameter}       & \textbf{Value}   & \textbf{Hyperparameter} & \textbf{Value} \\
\midrule
\multicolumn{4}{l}{\textbf{Training}} \\ \midrule
Learning rate                 & $1 \times 10^{-4}$  & Target network soft-update rate    & 0.05       \\
Batch size                    & 1024 &
Buffer size                   & 1\_000\_000\\   Optimizer                     & Adam  &
Sampling                      & Uniform\\ Discount factor ($\gamma$)  & 0.99 & Num of vector envs & 10 (Only in DMC tasks) \\
Sample batch size & 20 \\
\midrule
\multicolumn{4}{l}{\textbf{World Model learning}} \\ \midrule
Dynamics loss coefficient ($\beta_1$)   & 0.5 & Representation loss coefficient ($\beta_2$)  & 0.1 \\
Learning reate  & $1 \times 10^{-4}$ & Optimizer                     & Adam  \\
\midrule
\multicolumn{4}{l}{\textbf{Diffusion model}} \\ \midrule
Diffusion steps     & 100 & Noise Schedule  & Cosine \\
Guidance scale     & 1.0 & $\epsilon$-prediction  & True  \\
Denoising network     & [256,256,256] & Activation & GELU \\
Optimizer  & Adam &Learning rate &$1 \times 10^{-5}$ \\
\midrule
\multicolumn{4}{l}{\textbf{Actor}} \\ \midrule
Minimum policy log std   & -20 & Policcy network &  [256,256,256]  \\
Maximum policy log std   & 0.5  & Activation in hidden dim&  GELU \\
Learning rate   & $1 \times 10^{-4}$ & Activation in output dim &  Linear   \\
\midrule
\multicolumn{4}{l}{\textbf{Critic}} \\ \midrule
Value network  & [256,256,256] & Activation in hidden dim   & GELU \\
Learning rate  & $1 \times 10^{-4}$ & Activation in output dim &  Linear \\
\midrule
\multicolumn{4}{l}{\textbf{Architecture (8M)}} \\ \midrule
Transformer layers                & 2 & Latent space dimension        & 256 \\
Transformer heads             & 8 & Dropout      & 0.1 \\
MLP activation                & GELU &  Normalization             & LayerNorm \\
\bottomrule
\end{tabular}
\end{small}
\end{center}
\vskip -0.1in
\end{table}
\section{Supplemental clarification}

\subsection{Design of One-step imagination world model}
\label{wm}

\begin{figure}[t]
  \centering
    \includegraphics[width=0.8\textwidth,trim={0.1cm 0.1cm 0.1cm 0.1cm},   clip]{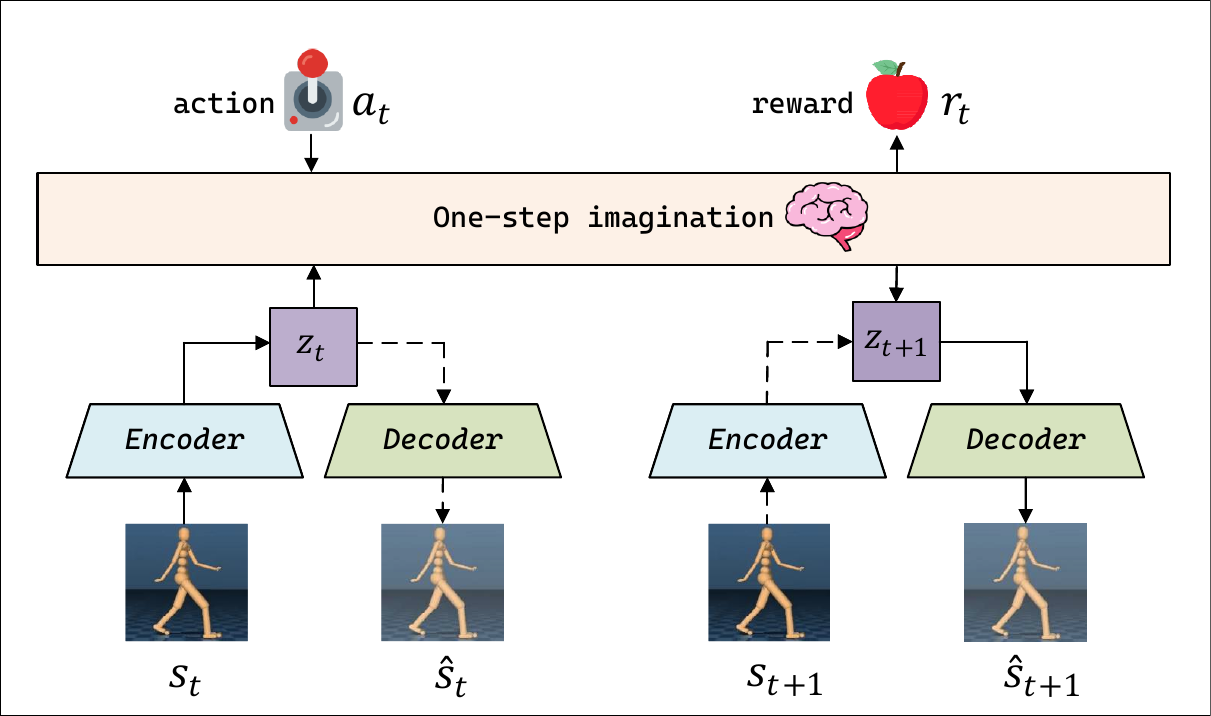}
  \caption{\textbf{One-step imagination world model.} During training, the current state $s_t$ is encoded by the representation network $h_\phi$ into a latent representation $\mathbf{z}_t$. Given this latent state and the action $a_t$, the one-step world model predicts the next latent state $\mathbf{z}_{t+1}$, the immediate reward $\hat{r}_t$, and termination factor $\hat{c}_t$. The solid-line process represents inference, while the dashed-line process is used for loss function construction.}
  \label{fig:wm}
\end{figure}

In MoGE, the structure of the one-step imagination world model is illustrated in Figure \ref{fig:wm}.
Since MoGE is developed under the MDP setting. The reason why modeling dynamics in a latent space instead of directly learning the mapping
$f(s,a)→s'$ is motivated by several practical considerations:

\textbf{(A))} Learning a compact latent representation helps capture abstract, task-relevant features of the environment, improving generalization and training efficiency even under MDPs. Besides, different dimensions of the state may contribute unevenly to dynamic modeling. Using an encoder to transform the raw state allows the model to extract the most relevant components, which facilitates more accurate transition prediction.

\textbf{(B)} Latent dynamics are often smoother and easier to predict compared to raw state transitions, especially in high-dimensional environments. Fitting dynamics in a latent space typically improves the accuracy of transition modeling.

\textbf{(C)} Decoupling representation learning from dynamics prediction allows better reuse and transfer of components. In fact, in MoGE, the policy network is built on top of the same encoder, enabling a unified state representation that facilitates more effective policy learning.

To further support our claim, a simple experiment is conducted across three environments (each with three random seeds) using the same diffusion and baseline algorithm setup. This experiment compares the TAR between vanilla dynamics (Using the Transformer as a predictor of next states and rewards) and latent dynamics. The results are demonstrated in the Table \ref{tab:mpge_ablation}:

\begin{table}[t]
\centering
\fontsize{9pt}{9}\selectfont
\setlength{\tabcolsep}{6pt}
\caption{Ablation of MoGE with vanilla vs. latent dynamics on humanoid control tasks.}
\begin{tabular}{lcc}
\toprule
Env & MoGE w/ latent dynamics & MoGE w/ vanilla dynamics \\
\midrule
Humanoid-run   & \textbf{489} $\pm$ 9   & 408 $\pm$ 35 \\
Humanoid-walk  & \textbf{892} $\pm$ 19  & 684 $\pm$ 71 \\
Humanoid-stand & \textbf{907} $\pm$ 7   & 785 $\pm$ 88 \\
\bottomrule
\end{tabular}
\label{tab:mpge_ablation}
\end{table}

The results show that, compared to directly mapping the current state and action to the next state, introducing a latent space leads to better learning of both environment transitions and state representations in the policy network.
\subsection{Broader baseline experiments with model-free algorithms}
To further validate the effectiveness of MoGE compared with active exploration, we perform extra experiments with more mainstream model-free algorithms. All the algorithms are evaluated in standard settings and tested on the 3 OpenAI Gym tasks: Walker2d-v3, Humanoid-v3, and Halfcheetah-v3.

\textbf{Deep Deterministic Policy Gradient (DDPG)} \cite{lillicrap2015continuous}: an off-policy actor-critic method that leverages deterministic policies and experience replay for efficient learning in continuous action spaces.

\textbf{Trust Region Policy Optimization (TRPO)} \cite{schulman2015trust}: an on-policy method that optimizes policies by enforcing a trust region constraint to ensure stable updates.

\textbf{Proximal Policy Optimization (PPO)} \cite{schulmanProximal2017}: improved upon TRPO by using a clipped surrogate objective for simpler and more efficient training.

\textbf{Soft Actor-Critic (SAC)} \cite{haarnoja2018soft}: introduces maximum entropy to encourage exploration and improve stability, making it well-suited for complex, high-dimensional tasks.

\begin{figure}[h]
  \centering
  \begin{subfigure}[b]{0.32\textwidth}
    \includegraphics[width=\textwidth,trim={0.18cm 0.15cm 0.18cm 0.15cm},   clip]{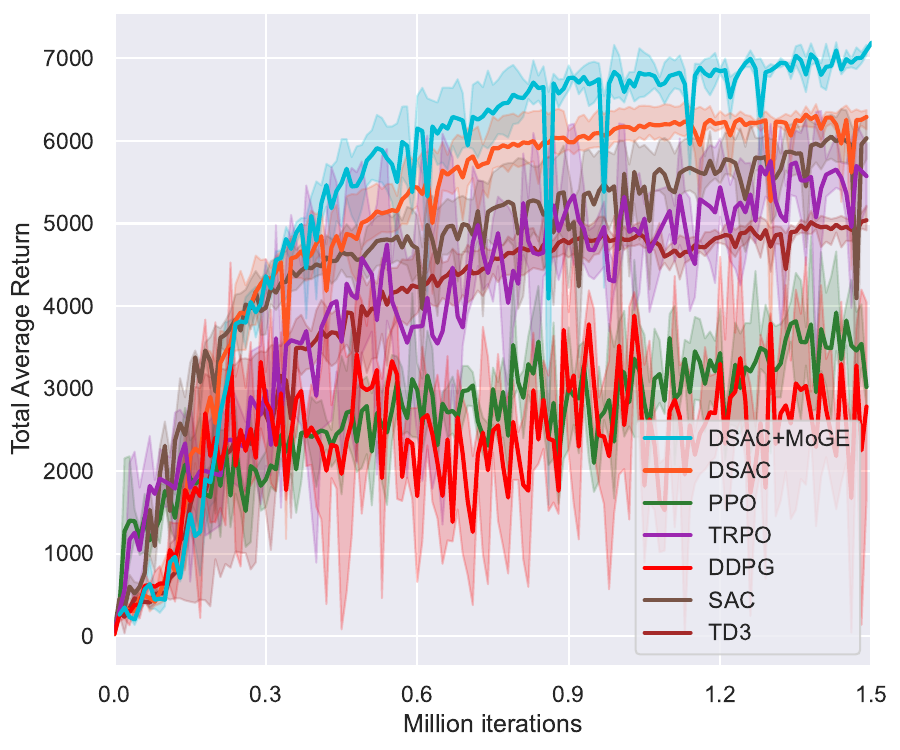}
    \captionsetup{width=0.75\linewidth}
    \caption{Walker2d-v3}
    \label{fig:sup_walker2d}
  \end{subfigure}
  \begin{subfigure}[b]{0.32\textwidth}
    \includegraphics[width=\textwidth,trim={0.18cm 0.15cm 0.18cm 0.15cm},   clip]{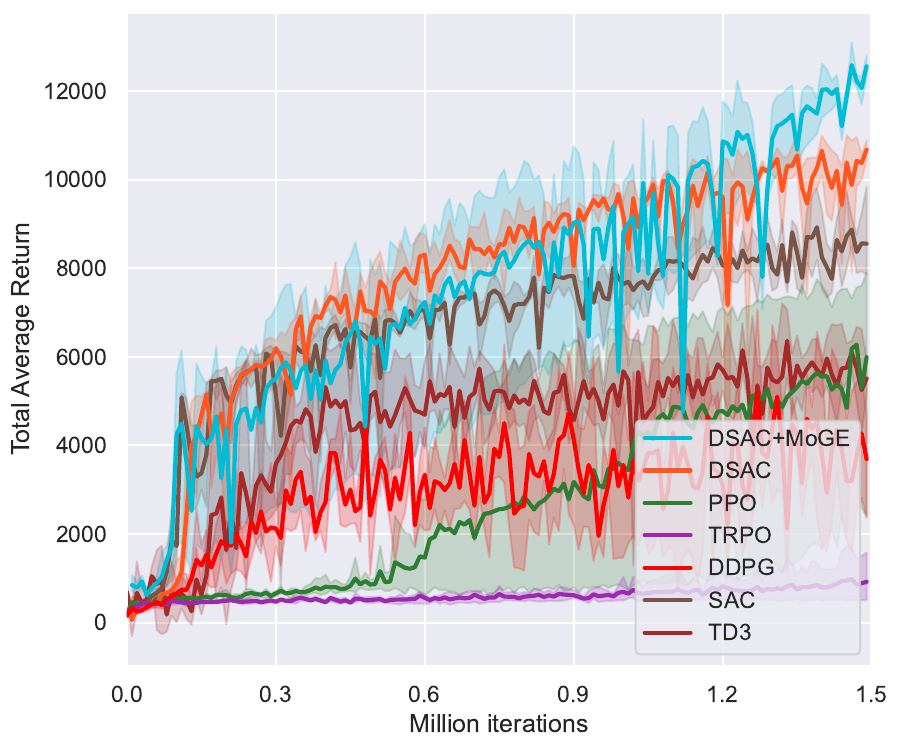}
    \captionsetup{width=0.75\linewidth}
    \caption{Humanoid-v3}
    \label{fig:sup_humanoid}
  \end{subfigure}
  \begin{subfigure}[b]{0.32\textwidth}
    \includegraphics[width=\textwidth,trim={0.18cm 0.15cm 0.18cm 0.15cm},   clip]{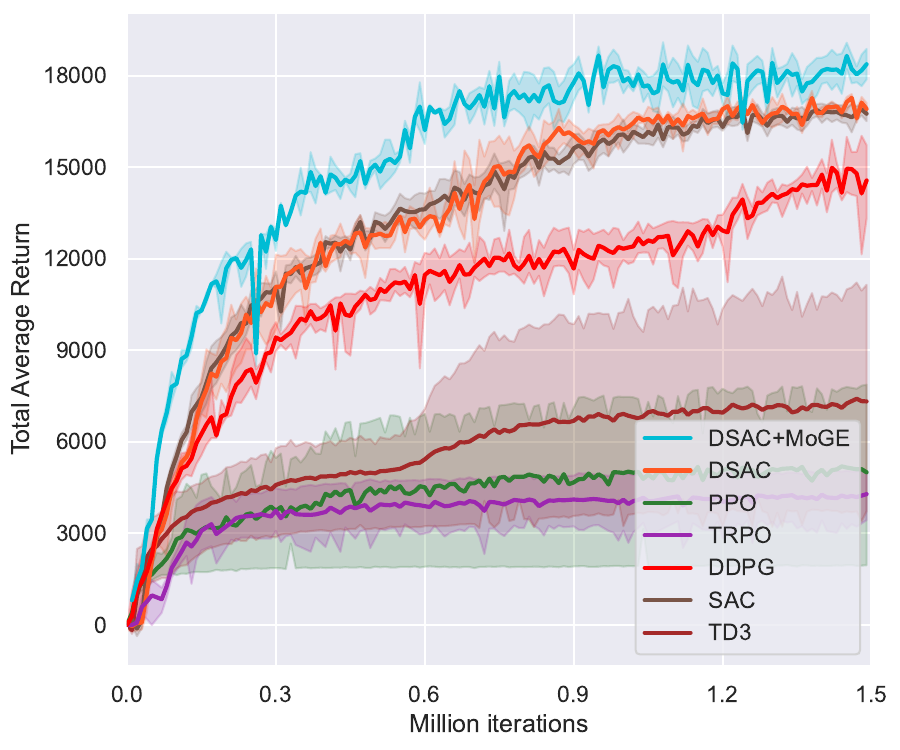}
    \captionsetup{width=0.75\linewidth}
    \caption{Halfcheetah-v3}
    \label{fig:sup_Halfcheetah}
  \end{subfigure}

  \caption{Supplemental study curves with 6 mainstream model-free algorithms.}
  \label{fig:sup_training_curves}
\end{figure}

All the training curves are illustrated in Figure \ref{fig:sup_training_curves} and the detailed results are listed in Table~\ref{tab:sup-mf}.

\begin{table}[h]
\centering
\fontsize{8pt}{10}\selectfont
\setlength{\tabcolsep}{2pt}
\caption{Total Average Return (TAR) on 3 OpenAI Gym tasks for supplemental experiments. Mean ± Std over 3 seeds. \textbf{Bold} = best,  Higher is better.}
\begin{tabular}{lcccccccccc}
\toprule
\textbf{Environment}
& TD3 & SAC & DDPG & TRPO & PPO & DSAC & DSAC+MoGE \\
\midrule
Walker2d-v3  & 5031.1 ± 84.2 & 5997.0 ± 291.1 & 3268.1 ± 240.9 & 5635.6 ± 211.9  & 3880.9 ± 327.7 & 6501.1 ± 87.3 & \textbf{6978.4 ± 68.7 }\\
Humanoid-v3    & 5967.1 ± 547.8 & 8831.7 ± 352.2 & 4548.7 ± 807.6 & 947.2 ± 503.9  & 6011.0 ± 2014.4  & 11004.0 ± 121.5 & \textbf{12151.1 ± 35.4} \\
Halfcheetah-v3    &  7363.0 ± 3666.4& 16921.3 ± 380.8 & 14793.1 ± 462.3 & 4207.4 ± 756.6  & 5139.2 ± 2392.8& 17324.7 ± 41.1 & \textbf{18054.9 ± 459.6} \\
\midrule
AVG.GYM      & 6120.4 ± 1432.8 & 10583.3 ± 341.4 & 7536.6 ± 503.6 & 3596.7 ± 490.8 & 5010.4 ± 1578.3 & 11472.7 ± 93.0 & \textbf{12394.8 ± 187.9} \\
\bottomrule
\end{tabular}
\label{tab:sup-mf}
\end{table}

\subsection{Broader baseline experiments with model-based algorithms}
Similarly, to validate the performance of MoGE when integrated with reinforcement learning algorithms, we additionally select two mainstream model-based RL algorithms for comparison. These algorithms are sourced from the \url{https://github.com/nicklashansen/tdmpc2}, and the experiments are performed in DMC Humanoid tasks:

\textbf{TD-MPC2} \cite{hansenTDMPC22024}: a model-based reinforcement learning algorithm that combines temporal difference learning with model-predictive control to enhance sample efficiency and long-horizon planning.

\textbf{DreamerV3} \cite{hafner2023mastering}: a model-based reinforcement learning framework that leverages a world model for imagination-based training, enabling effective policy learning with high sample efficiency in continuous control tasks.

\begin{figure}[h]
  \centering
  \begin{subfigure}[b]{0.3\textwidth}
    \includegraphics[width=\textwidth,trim={0.18cm 0.15cm 0.18cm 0.15cm},   clip]{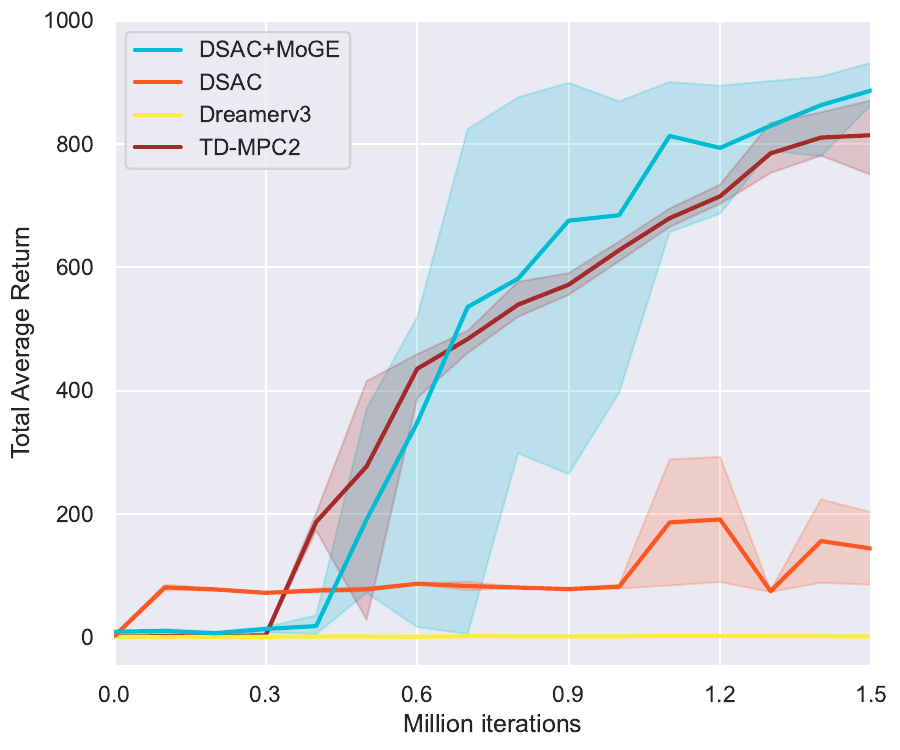}
    \captionsetup{width=0.75\linewidth}
    \caption{Humanoid-walk}
    \label{fig:sup_humanwalk}
  \end{subfigure}
  \begin{subfigure}[b]{0.3\textwidth}
    \includegraphics[width=\textwidth,trim={0.18cm 0.15cm 0.18cm 0.15cm},   clip]{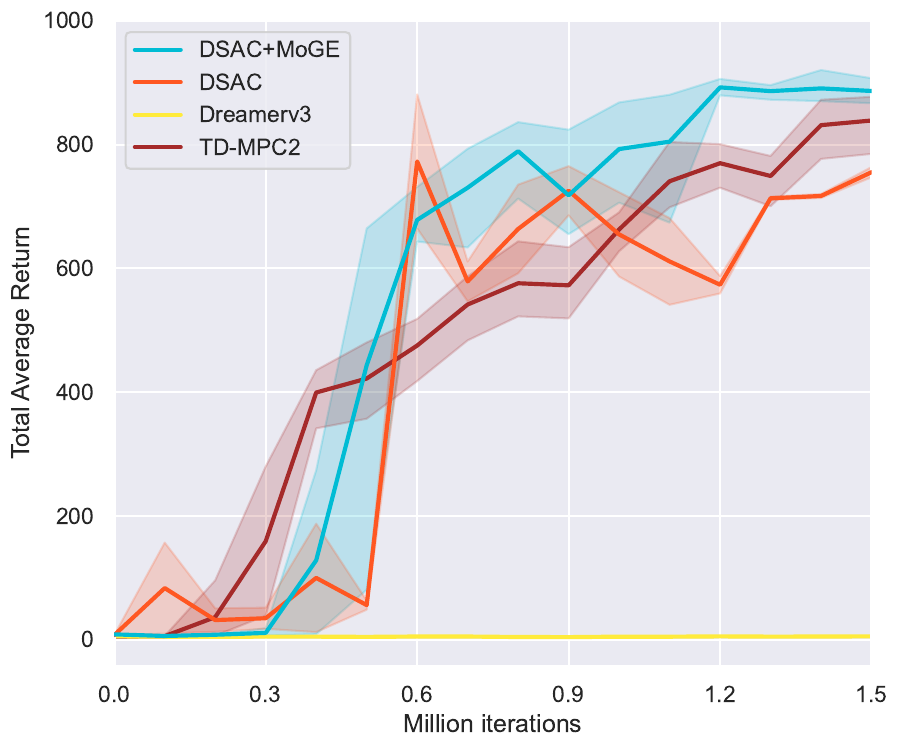}
    \captionsetup{width=0.75\linewidth}
    \caption{Humanoid-stand}
    \label{fig:sup_humanstand}
  \end{subfigure}
  \begin{subfigure}[b]{0.3\textwidth}
    \includegraphics[width=\textwidth,trim={0.18cm 0.15cm 0.18cm 0.15cm},   clip]{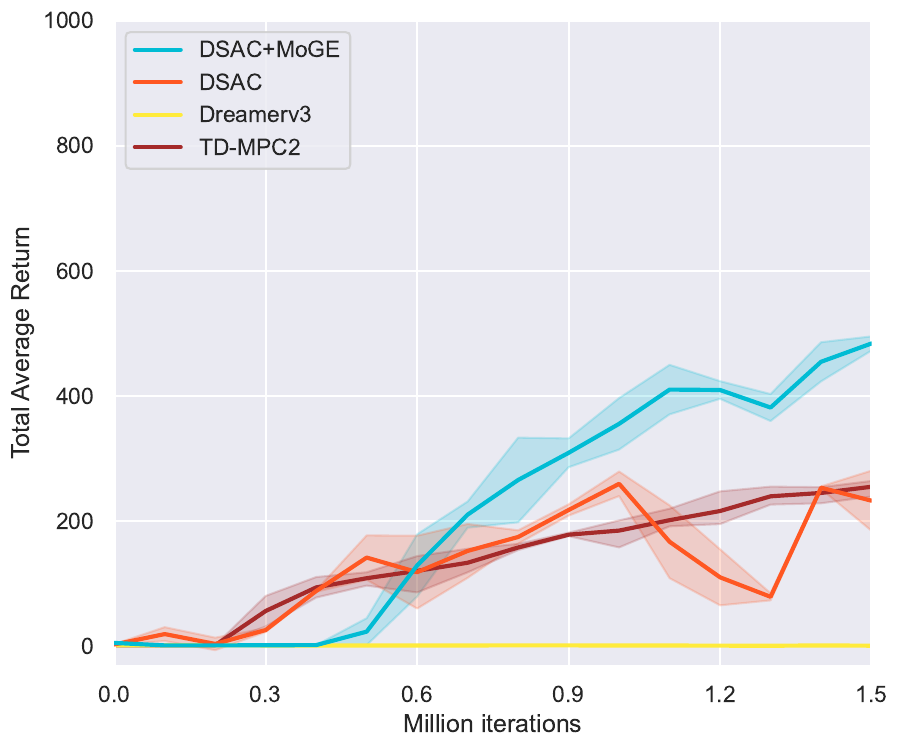}
    \captionsetup{width=0.75\linewidth}
    \caption{Humanoid-run}
    \label{fig:sup_humanrun}
  \end{subfigure}

  \caption{Supplemental study curves with 2 mainstream model-based algorithms.}
  \label{fig:sup_training_curves_mb}
\end{figure}

All the training curves are illustrated in Figure \ref{fig:sup_training_curves_mb} and the detailed results are listed in Table~\ref{tab:sup-mb}.
\begin{table}[h]
\centering
\fontsize{8pt}{10}\selectfont
\setlength{\tabcolsep}{4pt}
\caption{Total Average Return (TAR) on 3 DMC Suite tasks for supplemental experiments. Mean ± Std over 3 seeds. \textbf{Bold} = best,  Higher is better.}
\begin{tabular}{lcccccccccc}
\toprule
\textbf{Environment}
& DSAC & DreamerV3 & TD-MPC2 & DSAC+MoGE \\
\midrule
Humanoid-walk  & 146.5 ± 60.9 & 0.9 ± 0.4 & 814.3 ± 49.8& \textbf{891.7 ± 19.1}\\
Humanoid-stand    & 776.6 ± 15.6  & 5.6 ± 0.3 & 838.9 ± 39.2 & \textbf{907.5 ± 6.9}\\
Humanoid-run  &267.4 ± 3.9   &0.8 ± 0.4 & 254.6 ± 11.1 & \textbf{488.9 ± 8.7}  \\
\midrule
AVG.DMC      & 396.8 ± 26.8 & 2.4 ± 0.4 & 635.9 ± 33.4 & \textbf{762.7 ± 11.6} \\
\bottomrule
\end{tabular}
\label{tab:sup-mb}
\end{table}

\begin{table*}[t]
\centering
\setlength{\tabcolsep}{3pt}
\fontsize{6pt}{10}\selectfont
\begin{minipage}[t]{0.45\textwidth}
\vspace{0pt}
\centering
\caption{TAR for supplemental experiments with \textbf{modified AC methods}.}
\begin{tabular}{lccccc}
\toprule
Env & DSAC+MoGE & Simba & BRO & RedQ & DroQ \\
\midrule
Humanoid-run & \textbf{489} $\pm$ 9 & 268 $\pm$ 40 & 417 $\pm$ 15 & 187 $\pm$ 12 & 164 $\pm$ 21 \\
Humanoid-walk & \textbf{892} $\pm$ 19 & 801 $\pm$ 13 & 881 $\pm$ 25 & 665 $\pm$ 5 & 682 $\pm$ 14 \\
Humanoid-stand & 907 $\pm$ 7 & \textbf{920} $\pm$ 14 & 905 $\pm$ 3 & 902 $\pm$ 4 & 896 $\pm$ 6 \\
\bottomrule
\label{tab:modi}
\end{tabular}
\end{minipage}
\hfill
\begin{minipage}[t]{0.45\textwidth}
\vspace{0pt}
\centering
\caption{TAR for supplemental experiments with \textbf{enhanced exploration approaches}.}
\begin{tabular}{lcccc}
\toprule
Env & DSAC+MoGE & Plan2Explore & MaxInfoRL & OMBRL \\
\midrule
Humanoid-run & \textbf{489} $\pm$ 9 & 311 $\pm$ 12 & 197 $\pm$ 4 & 262 $\pm$ 13 \\
Humanoid-walk & \textbf{892} $\pm$ 19 & 588 $\pm$ 7 & 481 $\pm$ 7 & 678 $\pm$ 5 \\
Humanoid-stand & 907 $\pm$ 7 & 801 $\pm$ 14 & 844 $\pm$ 8 & 769 $\pm$ 11 \\
\bottomrule
\label{tab:explore}
\end{tabular}
\end{minipage}
\end{table*}

\subsection{Broader baseline experiments with modified Actor-Critic algorithms}
To provide a more comprehensive evaluation, some half-formulated methods that were modified based on the traditional Actor-Critic method can be included. We conduct the experiments in DMC Humanoid tasks, which are one of the hardest tasks within 3 seeds, and the introductions of the methods are illustrated as follows:

\textbf{REDQ} \cite{chen2021randomized}: a model-free RL algorithm that employs a large ensemble of Q-functions with randomized subset updates to improve sample efficiency and reduce overestimation bias.

\textbf{DroQ} \cite{hiraoka2021dropout}: a lightweight variant of ensemble Q-learning that applies dropout regularization to approximate uncertainty and achieve doubly efficient learning.

\textbf{BRO} \cite{nauman2024bigger}: a scalable actor-critic framework that leverages network scaling, regularization, and optimism to balance compute and sample efficiency in continuous control.

\textbf{Simba} \cite{lee2024simba}: a large-scale RL framework that exploits simplicity bias in over-parameterized networks to improve stability, generalization, and training efficiency.

As shown in Table \ref{tab:modi}, except for the simplest case (Humanoid-Stand), MoGE achieves the best performance across all more challenging tasks.
\subsection{Broader baseline experiments with enhanced exploration approaches}
We have conducted additional experiments to study the effect of different exploration strategies on the DSAC algorithm via ablation comparisons. Due to the development and implementation effort involved, and time constraints, we focus on three representative environments—the Humanoid tasks in DMC—for this evaluation. The three approaches are as follows:

\textbf{Plan2Explore} \cite{sekar2020planning}: a model-based exploration method that uses self-supervised world models to plan informative trajectories without external rewards.

\textbf{MaxInfoRL} \cite{sukhija2024maxinforl}: an exploration framework that maximizes mutual information between states, actions, and returns to encourage diverse and informative behaviors.

\textbf{OMBRL} \cite{sukhija2025optimism}: a model-based exploration algorithm that incorporates optimism through intrinsic reward shaping, enabling scalable and principled exploration.

Demonstrated in Table \ref{tab:explore}, the results show that MoGE still offers a significant advantage compared to these methods.
Intrinsic exploration methods encourage visiting novel or unpredictable states, but their signals (e.g., prediction error, uncertainty) are often task-agnostic and may lead to uninformative or misaligned exploration. In contrast, MoGE leverages task-aware utility functions (e.g., TD-error, entropy) to generate states that are directly aligned with policy improvement objectives**. Besides, unlike intrinsic reward methods that require careful reward balancing and affect the policy's optimization objective, MoGE decouples exploration from reward design, enabling more stable training without interfering with the task-specific learning signal.

\section{Limitation and Future Work}
\label{limit_future}
While MoGE demonstrates strong performance in enhancing exploration and improving sample efficiency in reinforcement learning, several limitations remain.
First, the generation of critical states through the diffusion-based generator introduces additional computational overhead compared to standard replay buffer sampling.
Second, MoGE assumes that the state distribution learned by the generator aligns well with the replay buffer's occupancy measure. In practice, minor discrepancies may arise.
Overall, while MoGE enhances exploration capabilities, its time cost is influenced by the quality of the learned state distribution and the smoothness of the conditional diffusion process.

In the future, we may explore integrating MoGE with on-policy RL frameworks to enable real-time generation of critical states, further enhancing exploration efficiency during live interactions. Additionally, investigating more expressive utility functions for the diffusion-based generator could improve the selection of high-value states, optimizing policy learning. Moreover, dynamically adjusting the generator's sampling strategy based on task complexity and training progress may further boost robustness and generalization.

\section{Positive and Negative Social Impact}
\label{impact}
Our method, MoGE, performs exploration augmentation in reinforcement learning by generating critical samples through a diffusion-based generator and a world model, significantly improving sample efficiency and policy performance in complex control tasks. This capability has promising implications for real-world applications such as embodied AI, autonomous driving, and large-scale decision-making systems, where efficient exploration of vast state spaces is crucial.  However, the ability to synthetically generate exploration samples may lead to overconfidence in simulation-trained policies, increasing risks if prematurely deployed in real-world environments. We advocate for careful validation and ethical considerations to ensure the responsible and safe application of MoGE.

\newpage
\section*{NeurIPS Paper Checklist}

\begin{enumerate}

\item {\bf Claims}
    \item[] Question: Do the main claims made in the abstract and introduction accurately reflect the paper's contributions and scope?
    \item[] Answer: \answerYes{}
    \item[] Justification: We emphasize our core contributions in the abstract and introduction, highlighting the critical state generator, the one-step imagination world model, and the proposed RL training framework combined with MoGE.

\item {\bf Limitations}
    \item[] Question: Does the paper discuss the limitations of the work performed by the authors?
    \item[] Answer: \answerYes{}
    \item[] Justification: In Appendix \ref{limit_future} we discuss that the limitation of this work is the problem of time cost and the bias in critical state generation.

\item {\bf Theory assumptions and proofs}
    \item[] Question: For each theoretical result, does the paper provide the full set of assumptions and a complete (and correct) proof?
    \item[] Answer: \answerYes{}
    \item[] Justification: We provide complete assumptions in the formulation of the theorem and a complete proof of the theorem in Appendix \ref{app:thm}.

    \item {\bf Experimental result reproducibility}
    \item[] Question: Does the paper fully disclose all the information needed to reproduce the main experimental results of the paper to the extent that it affects the main claims and/or conclusions of the paper (regardless of whether the code and data are provided or not)?
    \item[] Answer: \answerYes{}
    \item[] Justification: We give details of the hyperparameters required for the experiments in the Appendix \ref{app: details}.

\item {\bf Open access to data and code}
    \item[] Question: Does the paper provide open access to the data and code, with sufficient instructions to faithfully reproduce the main experimental results, as described in supplemental material?
    \item[] Answer: \answerNo{}
    \item[] Justification: We will make the full code publicly available when the paper is accepted, but the core code is public with open access in Appendix \ref{app: details}.

\item {\bf Experimental setting/details}
    \item[] Question: Does the paper specify all the training and test details (e.g., data splits, hyperparameters, how they were chosen, type of optimizer, etc.) necessary to understand the results?
    \item[] Answer: \answerYes{}
    \item[] Justification: In the appendix section, we give a very detailed selection of all algorithmic hyperparameters, optimizers.

\item {\bf Experiment statistical significance}
    \item[] Question: Does the paper report error bars suitably and correctly defined or other appropriate information about the statistical significance of the experiments?
    \item[] Answer: \answerYes{}
    \item[] Justification: Whether the data is in a table or a picture, we report the error bars.

\item {\bf Experiments compute resources}
    \item[] Question: For each experiment, does the paper provide sufficient information on the computer resources (type of compute workers, memory, time of execution) needed to reproduce the experiments?
    \item[] Answer: \answerYes{}
    \item[] Justification: In the Evaluation Setups in Appendix \ref{app: details}, we give the CPU and GPU servers used for the computation and the time used to train 1.5 million using Humanoid-run as an example.

\item {\bf Code of ethics}
    \item[] Question: Does the research conducted in the paper conform, in every respect, with the NeurIPS Code of Ethics \url{https://neurips.cc/public/EthicsGuidelines}?
    \item[] Answer: \answerYes{}
    \item[] Justification: Yes.

\item {\bf Broader impacts}
    \item[] Question: Does the paper discuss both potential positive societal impacts and negative societal impacts of the work performed?
    \item[] Answer: \answerYes{}
    \item[] Justification: We discuss both the potential positive societal impacts and negative societal impacts of the work in Appendix \ref{impact}

\item {\bf Safeguards}
    \item[] Question: Does the paper describe safeguards that have been put in place for responsible release of data or models that have a high risk for misuse (e.g., pretrained language models, image generators, or scraped datasets)?
    \item[] Answer: \answerNA{}
    \item[] Justification:

\item {\bf Licenses for existing assets}
    \item[] Question: Are the creators or original owners of assets (e.g., code, data, models), used in the paper, properly credited and are the license and terms of use explicitly mentioned and properly respected?
    \item[] Answer: \answerYes{}
    \item[] Justification: All our baseline algorithms are evaluated under standard settings, and all open-source implementations are properly cited.

\item {\bf New assets}
    \item[] Question: Are new assets introduced in the paper well documented and is the documentation provided alongside the assets?
    \item[] Answer: \answerNA{}
    \item[] Justification:

\item {\bf Crowdsourcing and research with human subjects}
    \item[] Question: For crowdsourcing experiments and research with human subjects, does the paper include the full text of instructions given to participants and screenshots, if applicable, as well as details about compensation (if any)?
    \item[] Answer: \answerNA{}
    \item[] Justification:

\item {\bf Institutional review board (IRB) approvals or equivalent for research with human subjects}
    \item[] Question: Does the paper describe potential risks incurred by study participants, whether such risks were disclosed to the subjects, and whether Institutional Review Board (IRB) approvals (or an equivalent approval/review based on the requirements of your country or institution) were obtained?
    \item[] Answer: \answerNA{}
    \item[] Justification:

\item {\bf Declaration of LLM usage}
    \item[] Question: Does the paper describe the usage of LLMs if it is an important, original, or non-standard component of the core methods in this research? Note that if the LLM is used only for writing, editing, or formatting purposes and does not impact the core methodology, scientific rigorousness, or originality of the research, declaration is not required.

    \item[] Answer: \answerNA{}
    \item[] Justification:

\end{enumerate}

\end{document}